\documentclass[lettersize,journal]{IEEEtran}
\usepackage{amsmath,amsfonts}
\usepackage{algorithmic}
\usepackage{algorithm}
\usepackage{array}
\usepackage[caption=false,font=footnotesize]{subfig}
\usepackage{textcomp}
\usepackage{stfloats}
\usepackage{url}
\usepackage{verbatim}
\usepackage{graphicx}
\usepackage{pifont} % 用于 \ding{51} 和 \ding{55}

\newcommand{\cmark}{\ding{51}}
\newcommand{\xmark}{\ding{55}}
\usepackage{cite}
\usepackage{booktabs} % 用于三线表的 toprule, midrule, bottomrule, cmidrule
\usepackage{multirow} 
\usepackage[colorlinks=true, linkcolor=blue, citecolor=blue, urlcolor=blue]{hyperref}
\begin{document}

\title{GeoSearcher: Anchor-Guided Progressive Reasoning for Remote Sensing Visual Grounding with Process Supervision}

\author{Dianyu Wang, Peirong Zhang, Xuyang Li, Xiaoxuan Liu,~\IEEEmembership{Member,~IEEE}, and Lei Wang,~\IEEEmembership{Member,~IEEE}% \thanks{Dianyu Wang, Yidan Zhang, Peirong Zhang, Xuyang Li, Xiaoxuan Liu, and Lei Wang are with the School of XXX, XXX University, City, China.} \thanks{Corresponding author: Lei Wang.} }
        % <-this % stops a space
\thanks{This work was supported in part by Key Program of Chinese Academy of Sciences under Grant KGFZD-145-25-38 and RCJJ-145-24-13; Science and Disruptive Technology Program under Grant AIRCAS2024-AIRCAS-SDTP-03. (Corresponding authors: Lei Wang.)}% <-this % stops a space
\thanks{Dianyu Wang, Peirong Zhang, Xuyang Li and Lei Wang are with the Aerospace Information Research Institute, the Key Laboratory of Network Information System Technology (NIST), and the Key Laboratory of Target Cognition and Application Technology (TCAT), Chinese Academy of Sciences, Beijing 100190, China, and also with the School of Electronic, Electrical and Communication Engineering, University of Chinese Academy of Sciences, Beijing 100190, China (e-mail: wangdianyu25@mails.ucas.ac.cn;  zhangpeirong22@mails.ucas.ac.cn; lixuyang23@\allowbreak mails.ucas.ac.cn; wanglei002931@aircas.ac.cn).} \thanks{Xiaoxuan Liu are with the Aerospace Information Research Institute, the Key Laboratory of Network Information System Technology (NIST), and the Key Laboratory of Target Cognition and Application Technology (TCAT), Chinese Academy of Sciences, Beijing 100190, China (e-mail: liuxiaoxuan@aircas.ac.cn).} }

\markboth{}%
% \markboth{}%
{Shell \MakeLowercase{\textit{et al.}}: A Sample Article Using IEEEtran.cls for IEEE Journals}


\maketitle

\begin{abstract}
Recent multimodal large language models (MLLMs) have shown strong cross-modal understanding and coordinate generation abilities in visual grounding. However, transferring these abilities to remote sensing visual grounding (RSVG) remains challenging. High-resolution remote sensing images usually cover large-scale scenes, where targets are often extremely small and surrounded by numerous visually similar distractors. Meanwhile, queries often contain multiple clues, such as reference objects, spatial relations, and target attributes. Existing MLLM-based methods usually formulate RSVG as one-step coordinate generation, which may lead to unstable predictions for small-object localization and complex queries. To address these challenges, we propose GeoSearcher, which reformulates RSVG as an anchor-guided progressive reasoning process and realizes it through two coupled stages: Anchor-Centric Reasoning Supervised Fine-Tuning (ACR-SFT) and Process-Faithful Group Relative Policy Optimization (PF-GRPO). In ACR-SFT, anchor-centric reasoning data are used to teach the model to represent key visual clues as anchors and progressively integrate location, relational, and attribute clues around them. In PF-GRPO, Process-Aware Reward (PAR) and Reasoning-Informative Sample Selector (RISS) further optimize this reasoning behavior by jointly evaluating key reasoning steps and target localization, while focusing training on samples that are more beneficial for improving progressive reasoning. Through this design, GeoSearcher transforms large-scale visual search into a more constrained local reasoning process. Extensive experiments on DIOR-RSVG, OPT-RSVG, and VRS-Bench show that GeoSearcher outperforms existing state-of-the-art methods. The project will be released at https://github.com/wangdianyu954-xixi/GeoSearcher.
\end{abstract}

\begin{IEEEkeywords}
Remote sensing visual grounding, reasoning model, multimodal large language models, reinforcement learning.
\end{IEEEkeywords}

\section{Introduction}
\label{sec:intro}

\IEEEPARstart{R}{emote} sensing visual grounding (RSVG) aims to localize a specific object in satellite or aerial images according to a natural language description. By allowing users to specify targets with flexible language expressions, RSVG provides an intuitive interface for object localization and retrieval in complex remote sensing scenes. This task is important for a wide range of applications, such as disaster response, agricultural monitoring, and urban management.

\begin{figure}[t]
    \centering
    
    \includegraphics[width=\columnwidth,trim=0 10 0 0, clip]{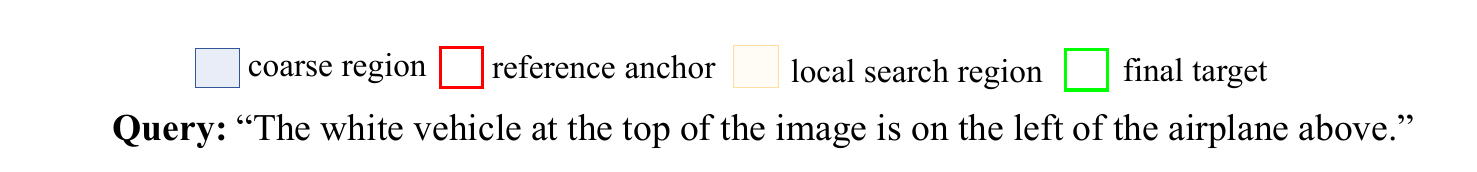}
    
    \vspace{0em}
    
    \subfloat[One-shot grounding.\label{fig:teaser1_direct}]{
        \includegraphics[width=0.42\columnwidth ,trim=30 15 100 0, clip]{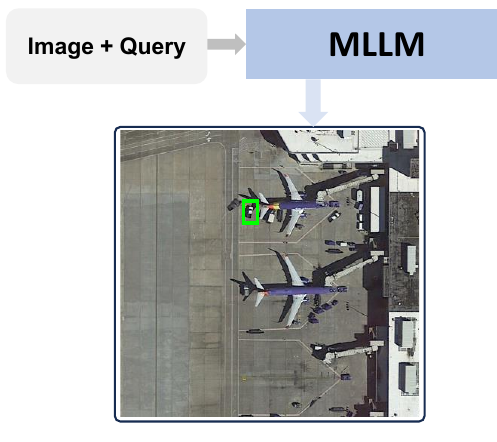}
    }
    \hfill
    \subfloat[Progressive reasoning.\label{fig:teaser1_agpi}]{
        \includegraphics[width=0.52\columnwidth ,trim=26 0 0 0, clip]{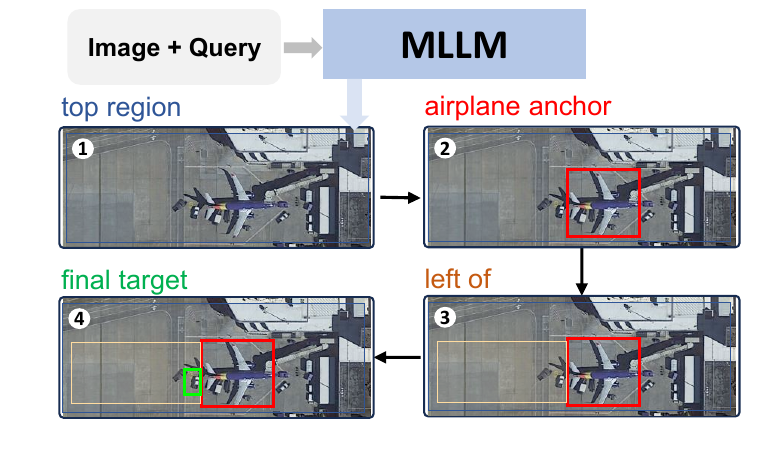}
    }
    
    \caption{Comparison between one-shot grounding and anchor-guided progressive reasoning in RSVG. (a) one-shot grounding predicts the target from the full image in one step, (b) whereas anchor-guided progressive reasoning progressively exploits multi-clues.}
    \label{fig:teaser1}
    % \vspace{-0.8em}
\end{figure}

Recent multimodal large language models (MLLMs)~\cite{bai2023qwenvlversatilevisionlanguagemodel},\cite{chen2023minigptv2largelanguagemodel},\cite{liu2023visual} have shown strong capability in visual grounding by unifying visual perception and language understanding within a generative framework.
Recently, many studies have further adapted MLLMs to RSVG~\cite{kuckreja2024geochat},\cite{zhou2024geoground},\cite{luo2024skysensegpt}.
As illustrated in Fig.~\ref{fig:teaser1_direct}, most existing MLLM-based RSVG methods formulate localization as \emph{one-shot coordinate generation}, where the model jointly encodes the query and the full image, and directly predicts the target object bounding box in a single step.
This formulation is effective when the target object is relatively large and can be identified from a few direct clues, but it becomes increasingly fragile in remote sensing scenes.

\begin{figure}[t!]
\centering
\includegraphics[width=\columnwidth]{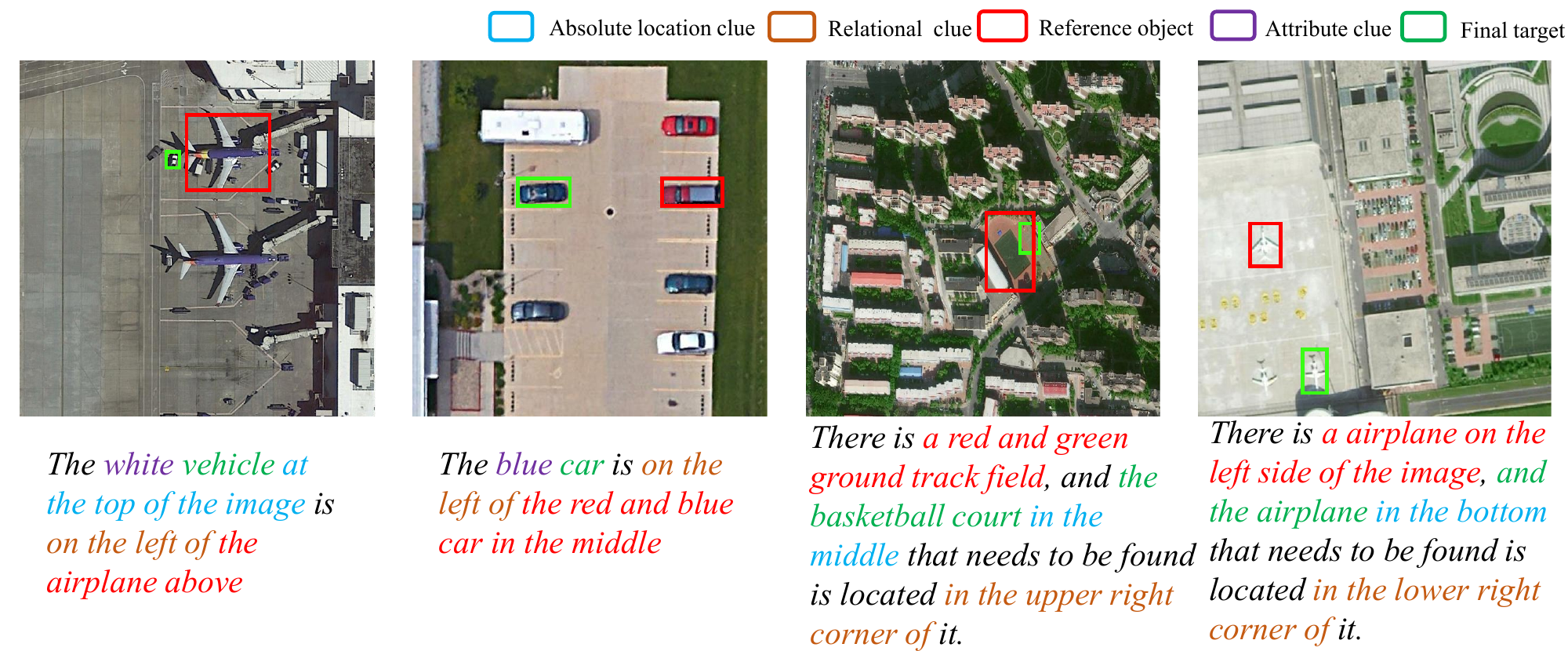}
\caption{Examples of complex RSVG. In high-resolution overhead scenes, a small target object often needs to be identified by combining a reference object with absolute, relational, and attribute clues.}
\label{fig:motivation}
% \vspace{-0.8em}
\end{figure}

As shown in Fig.~\ref{fig:motivation}, remote sensing images usually cover large geographic areas, while the referred object often occupies only a very small region.
Moreover, such scenes commonly contain many objects of the same category with complex spatial layouts\cite{li2024language}.
Therefore, to uniquely identify a small object, a query often needs to use a relatively salient object as a reference, such as ``The white vehicle at the top of image is on the left of the airplane above''.
This is similar to how people describe their own location by first referring to a nearby landmark.
Meanwhile, the query may also involve absolute location, relative position, scale comparison, and attribute description clues.
Under such conditions, the model must not only localize a tiny target object within a large visual search space, but also parse and integrate multiple clues.
\textbf{These observations motivate a reformulation of RSVG from one-shot coordinate generation to multi-step reasoning.}
Through multi-step reasoning, the model can first identify the reference object and exploit relevant clues step by step before final localization. 
This alleviates the burden of directly predicting the target bounding boxes in a single step from high-resolution imagery and complex query semantics.

Recent studies on reasoning-oriented large language models (LLMs) ~\cite{jaech2024openai},~\cite{guo2025deepseek} have shown that reinforcement learning (RL) based post-training can effectively elicit and strengthen multi-step reasoning abilities. Inspired by this paradigm, several vision-language studies in both natural and remote sensing scenarios have introduced RL based post-training for downstream tasks~\cite{fiaz2025geovlm},\cite{shen2025vlm},\cite{yao2026remotereasoner}. These methods usually optimize sampled responses with task-specific rewards, encouraging the model to generate more structured reasoning processes rather than relying only on direct answer prediction. Among them, Group Relative Policy Optimization (GRPO)\cite{shao2024deepseekmath} based methods have become a representative choice because they can achieve effective optimization with compact rule-based rewards~\cite{huang2026rsground,bai2025univg}. These studies further support the value of multi-step reasoning for complex vision-language tasks.

However, existing reasoning methods in the remote sensing domain still face limitations when handling complex RSVG tasks.
1) They usually lack explicit constraints that reliably associate the key reference object mentioned during reasoning with the correct image regions. Once this key object is not genuinely or correctly grounded, the error may propagate to subsequent reasoning and degrade final localization accuracy. The analysis in Section~\ref{subsec:anchor_quality} further supports this viewpoint.
2) Current GRPO-based optimization for visual grounding usually focuses on improving the final localization accuracy, with rewards mostly designed around the target object coordinates\cite{AyboraKksal2025TinyRSR1CV},\cite{wang2025geozero}. As a result, the reasoning process itself is only weakly supervised. Such sparse supervision may encourage shortcut learning, causing the model to bypass critical reasoning sections that should have been performed.

To address the fragility of one-shot coordinate generation in complex remote sensing scenes and the limitations of directly transferring multi-step reasoning paradigms to RSVG, we exploit the natural bridging role of reference object clues in uniquely identifying small objects. 
For complex queries, we wrap the verified reference object coordinates with special tokens \texttt{<ref\_bbox>} and treat the marked coordinate fragment as an anchor in the reasoning chain. 
This design establishes an explicit coordinate-level association between the reference object mention and its corresponding image region. 
Based on this formulation, \textbf{we reformulate RSVG from one-shot coordinate generation into an anchor-guided progressive reasoning problem}. We propose \textbf{GeoSearcher}, an \textbf{Anchor-Guided Progressive Reasoning Post-Training framework} for reasoning-oriented RSVG.
As shown in Fig.~\ref{fig:teaser1_agpi}, GeoSearcher follows an anchor-guided progressive reasoning strategy: it exploits coarse location clues, grounds the reference object, narrows the relevant local search region, and then localizes the final target object. For simple queries, GeoSearcher jointly uses the available clues for localization without enforcing unnecessary intermediate reasoning.

The GeoSearcher consists of two stages.
Specifically, Stage 1, \textbf{Anchor-Centric Reasoning Supervised Fine-Tuning (ACR-SFT)}, uses both progressive reasoning data with anchors and reasoning data without anchors, enabling the model to adaptively select an appropriate reasoning mode according to the query structure. Stage 2, \textbf{Process-Faithful Group Relative Policy Optimization (PF-GRPO)}, further aligns the reasoning process, where the \textbf{Process-Aware Reward (PAR)} jointly evaluates the key intermediate reasoning process and final object localization.
To make PF-GRPO more effective in improving progressive reasoning, we further consider the learning value of different samples at each training step. RSVG samples vary substantially in query complexity and image scene difficulty, so not all samples contribute equally to improving the model. For example, for some samples, the required reasoning ability has already been largely acquired by the model at the current stage, so further optimizing the model with these samples provides limited incremental value for improving progressive reasoning. Moreover, as the model gradually improves during training, the learning value of each sample for progressive reasoning is not fixed. 
PAR can evaluate how well the current model performs progressive reasoning on a given sample, but this signal is only available after rollout generation. To address this issue, we introduce the \textbf{Reasoning-Informative Sample Selector (RISS)}, which is updated online along with the model to estimate the expected PAR response of candidate samples during data sampling. RISS prioritizes samples that are expected to provide more informative supervision at the current training stage.

Extensive experiments on DIOR-RSVG~\cite{zhan2023rsvg}, OPT-RSVG~\cite{li2024language}, and VRS-Bench~\cite{li2024vrsbench} show that GeoSearcher consistently outperforms existing state-of-the-art methods, especially in small object scenarios where localization is driven by multi-clue reasoning.

The main contributions of this work are summarized as follows:
\begin{itemize}
    \item We reformulate RSVG as anchor-guided progressive reasoning. Under this formulation, the model progressively narrows the search region around anchors through multi-clue integration before localizing small target objects.

    \item We propose GeoSearcher, a two-stage post-training framework composed of ACR-SFT and PF-GRPO. GeoSearcher guides the MLLM to perform query-adaptive grounding. For complex queries, it localizes targets through progressive reasoning; for simple queries, it avoids unnecessary reasoning.

    \item We develop a pipeline to synthesize Anchor-Centric Reasoning data for ACR-SFT, where reference object mentions are explicitly associated with correct image regions. We design PAR and RISS to improve the faithfulness and efficiency of reasoning process optimization during PF-GRPO.

    \item Extensive experiments on DIOR-RSVG, OPT-RSVG, and VRS-Bench demonstrate that GeoSearcher achieves state-of-the-art performance and improves the faithfulness of the reasoning process.
\end{itemize}

\section{Related work}
% In this section, we review the research most relevant to our study. 
% We first discuss remote sensing visual grounding in Section~\ref{subsec:rw_rsvg}. We then review recent reasoning-based RSVG methods in Section~\ref{subsec:rw_reasoning}. Finally, we summarize reinforcement learning methods for MLLMs in Section~\ref{subsec:rw_rl}.
\subsection{Remote Sensing Visual Grounding}
\label{subsec:rw_rsvg}
Early studies mainly relied on task-specific discriminative frameworks
\cite{sun2022visual,li2024language,zhan2023rsvg,lan2024language}, where cross-modal representation learning, attention interaction, and coordinate regression were jointly designed for the grounding task. These methods established an important foundation for RSVG, but their modeling capacity was still largely constrained by specialized architectures and implicit cross-modal matching.

With the rapid development of MLLMs \cite{bai2023qwenvlversatilevisionlanguagemodel,chen2023minigptv2largelanguagemodel,liu2023visual}, visual grounding has shifted from discriminative prediction to a more unified generative paradigm. In the general domain, representative works such as Shikra \cite{chen2023shikra}, Ferret \cite{you2023ferret}, and Kosmos-2 \cite{peng2023kosmos} showed that grounding can be integrated into autoregressive generation through numerical coordinates, special markers, or grounding-aware token designs. These advances provided methodological inspiration for RSVG.

Building on this trend, a growing number of studies have adapted MLLMs to the remote sensing domain. GeoChat \cite{kuckreja2024geochat}, LHRS-Bot \cite{muhtar2024lhrs}, H2RSVLM \cite{pang2024h2rsvlm}, and SkySenseGPT \cite{luo2024skysensegpt} improved remote sensing instruction following through large-scale data construction and unified multi-task learning, enabling captioning, visual question answering (VQA), and grounding within a single framework. EarthGPT \cite{zhang2024earthgpt} further extended this paradigm to multi-sensor remote sensing imagery. In parallel, some works focused more directly on grounding-oriented spatial output modeling. For example, GeoGround \cite{zhou2024geoground} unified multiple grounding output formats, and GeoPixel \cite{shabbir2025geopixel} further extended grounding to pixel-level localization.
Despite these advances, most existing MLLM-based methods still formulate localization as direct coordinate generation. Such one-shot prediction remains challenging in remote sensing imagery. This motivates us to move beyond direct generation toward the multi-step grounding process.

\subsection{Reasoning-based RSVG}
\label{subsec:rw_reasoning}
Recent studies have introduced reasoning-guided paradigms into remote sensing vision-language tasks, providing useful foundations for reasoning-based RSVG. Instead of relying only on direct answer prediction, these methods encourage models to generate intermediate reasoning traces, making the decision process more interpretable and better suited to complex spatial layouts, small objects, and multi-clue descriptions in remote sensing scenes.

Some works explore how to enhance general remote sensing reasoning capability\cite{shao2025asking},\cite{hu2025ringmo}. TinyRS-R1 \cite{AyboraKksal2025TinyRSR1CV} adopts a multi-stage pipeline with reasoning supervision and GRPO alignment, showing that reasoning-oriented training can improve scene understanding and grounding for MLLM. RSThinker \cite{liu2025towards} formulates remote sensing analysis as a perceptually grounded multi-step reasoning process and improves factual correctness through staged reasoning alignment.

Beyond general remote sensing reasoning, another line of work focuses on reasoning-oriented training for region-level referring tasks. RemoteReasoner~\cite{yao2026remotereasoner} and Geo-R1~\cite{zhang2025geo} investigate how explicit reasoning can benefit geospatial referring tasks that require both language understanding and region-level visual localization. More directly related to RSVG, RSGround-R1 \cite{huang2026rsground} introduces synthetic reasoning data together with spatially guided optimization to improve position-aware grounding. These methods indicate that explicit reasoning can benefit complex remote sensing queries. However, existing reasoning-based RSVG methods still represent intermediate reasoning mainly as textual trajectories. 
Critical visual clues in the reasoning, especially reference objects, are not reliably grounded to image regions. Even if a related region is produced, its accuracy is often not explicitly constrained. Our approach links reference mentions to image regions and explicitly constrains the correctness of these associations.

\subsection{Reinforcement Learning for MLLMs}
\label{subsec:rw_rl}

RL has recently become an important post-training strategy for improving the reasoning capability of MLLMs. 
Representative post-training paradigms, including DPO~\cite{rafailov2023direct}, PPO~\cite{schulman2017proximal}, and GRPO~\cite{shao2024deepseekmath}, have been widely used to align model outputs with preference signals, task objectives, or rule-based rewards. 
In particular, GRPO has been widely adopted because it enables effective optimization with relatively simple rule-based rewards. 
Beyond general multimodal reasoning\cite{hong2025glm},\cite{team2025kimi}, recent studies have also extended RL-based post-training to grounding-related tasks~\cite{bai2025univg,shen2025vlm}, showing its potential for improving referring and localization behavior.

In the remote sensing domain, RL-based post-training has also been introduced into grounding-related models~\cite{yao2026remotereasoner,wang2025geozero,huang2026rsground}. 
Existing studies have shown that RL can improve remote sensing grounding performance. 
However, their reward designs are still mainly centered on the final grounding accuracy. 
This is broadly consistent with many RL settings in the general domain, where optimization is primarily driven by outcome-level supervision. 
However, for complex RSVG, final object correctness alone is often insufficient. 
A more critical issue is whether the intermediate reasoning process has been correctly executed. 
Our work focuses on aligning the intermediate  process, rather than optimizing only the target bounding box.

\section{THE PROPOSED METHOD}
\label{sec:method}
\begin{figure*}[!t]
\centering
\includegraphics[
    width=\textwidth,
    height=0.42\textheight,
    keepaspectratio,
    trim=0 45 0 0,
    clip
]{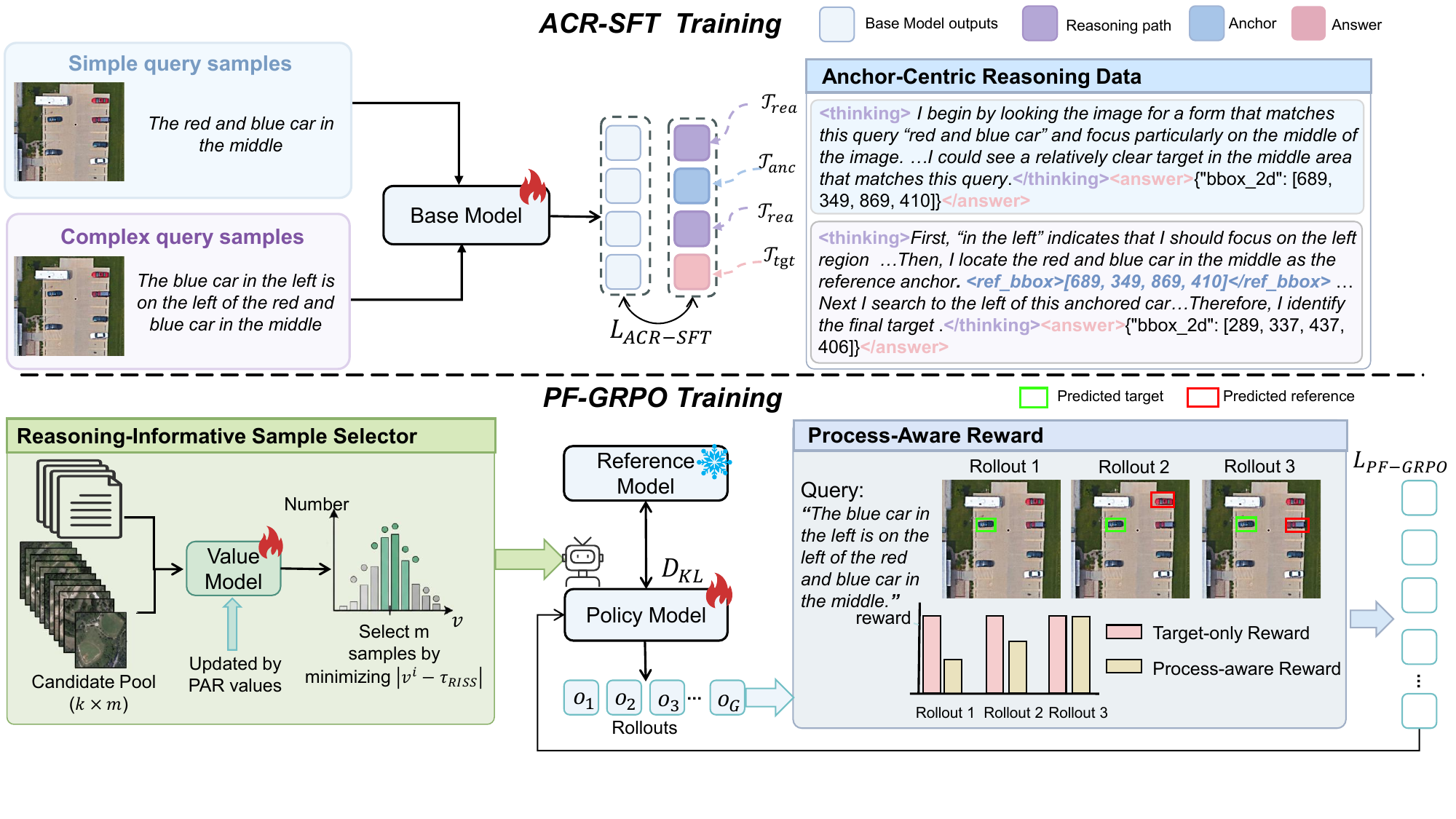}
\caption{Overall framework of GeoSearcher. In Stage 1, the base model is optimized by ACR-SFT on Anchor-Centric Reasoning data. In Stage 2, PF-GRPO further optimizes the reasoning process through two key designs: RISS and PAR.}
\label{fig:framework}
% \vspace{-0.8em}
\end{figure*}

This section presents the proposed GeoSearcher framework. As shown in Fig.~\ref{fig:framework}, the GeoSearcher consists of two stages. 
Stage 1, ACR-SFT, trains the model on both complex query samples with explicit anchors and simple query samples without anchors. 
This design encourages the model to adaptively choose the appropriate grounding mode according to the query structure: performing anchor-guided progressive reasoning for complex queries, while avoiding unnecessary reasoning detours for simple queries. 
Stage 2, PF-GRPO, further optimizes the reasoning process through RL. 
Its optimization is guided by PAR, which supervises key intermediate reasoning steps and target localization rather than only the target localization, and by RISS, which selects samples that are more informative for reasoning process optimization.
The remainder of this section is organized as follows.
Section~\ref{subsec:post_training} gives an overview of the two-stage training framework.
Section~\ref{subsec:data_pipeline} describes the construction and verification of anchor-centric reasoning data for ACR-SFT.
Section~\ref{subsec:par} and ~\ref{subsec:rids} introduce the PAR and RISS used in PF-GRPO.

\subsection{Anchor-Guided Progressive Reasoning Post-Training Framework}
\label{subsec:post_training}

\subsubsection{ACR-SFT}
\label{subsubsec:acr_sft}

In the first stage, we fine-tune the base model Qwen3VL-4B on Anchor-Centric Reasoning Data. 
The training set contains two types of samples: complex query samples with anchors $\mathcal{D}_{\mathrm{ref}}$, and simple query samples without inserted anchors $\mathcal{D}_{\mathrm{simp}}$.

We denote the training set as
\begin{equation}
\mathcal{D}_{\mathrm{acr}}
=
\mathcal{D}_{\mathrm{ref}}
\cup
\mathcal{D}_{\mathrm{simp}},
\end{equation}

For each image-query pair $x=(I,Q)$, let $z_Q\in\{0,1\}$ indicate whether the query requires an explicit reference object. 
The supervised output is defined as
\begin{equation}
y^{*}(x)
=
\begin{cases}
[r_{\mathrm{pre}}^{\mathrm{acr}}; A_{\mathrm{ref}}; r_{\mathrm{post}}^{\mathrm{acr}}; B_{\mathrm{tgt}}],
& z_Q=1, \\
[r_{\mathrm{simp}}^{\mathrm{acr}}; B_{\mathrm{tgt}}],
& z_Q=0,
\end{cases}
\end{equation}
where $[\cdot\,;\cdot]$ denotes sequence concatenation. 
$A_{\mathrm{ref}}=\texttt{<ref\_bbox>}B_{\mathrm{ref}}\texttt{</ref\_bbox>}$ is the anchor wrapped by special tokens, $B_{\mathrm{ref}}$ is the reference object coordinates, and $B_{\mathrm{tgt}}$ is the target object coordinates. 
$r_{\mathrm{pre}}^{\mathrm{acr}}$ and $r_{\mathrm{post}}^{\mathrm{acr}}$ denote the reasoning fragments before and after the anchor, respectively, and $r_{\mathrm{simp}}^{\mathrm{acr}}$ denotes the reasoning trajectory for simple queries.

The ACR-SFT objective is formulated as
\begin{equation}
\begin{aligned}
\mathcal{L}_{\mathrm{ACR\text{-}SFT}}
= -\mathbb{E}_{(x,y^{*})\sim\mathcal{D}_{\mathrm{acr}}}
\Bigg[
&\sum_{t\in\mathcal{T}_{\mathrm{rea}}}
\log p_{\eta}(y_t^{*}\mid x,y_{<t}^{*}) \\
&+
\sum_{t\in\mathcal{T}_{\mathrm{anc}}}
\log p_{\eta}(y_t^{*}\mid x,y_{<t}^{*}) \\
&+
\sum_{t\in\mathcal{T}_{\mathrm{tgt}}}
\log p_{\eta}(y_t^{*}\mid x,y_{<t}^{*})
\Bigg].
\end{aligned}
\end{equation}
where $\mathcal{T}_{\mathrm{rea}}$, $\mathcal{T}_{\mathrm{anc}}$, and $\mathcal{T}_{\mathrm{tgt}}$ denote the token positions of reasoning text, anchor, and target coordinates, respectively. 
For simple queries, $\mathcal{T}_{\mathrm{anc}}=\emptyset$. 

This objective enables the model to perform query-adaptive grounding. For complex queries, the model learns to use explicit reference anchors for progressive reasoning. For simple queries, it learns to avoid unnecessary anchor generation.

\begin{figure*}[!t]
    \centering
    \includegraphics[width=1\linewidth,trim=0 23 0 0, clip]{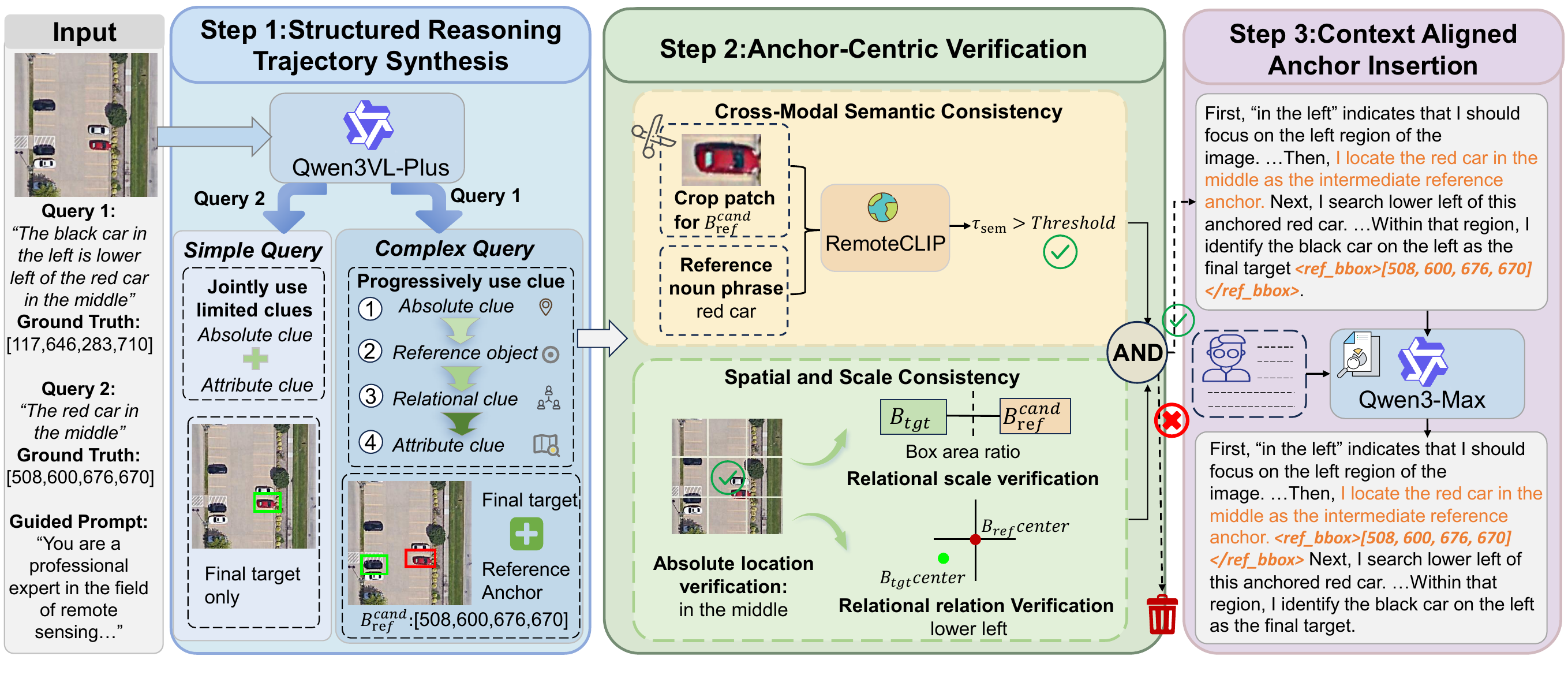}
    \caption{Pipeline of Anchor-Centric Reasoning Data construction for ACR-SFT. 1) synthesize trajectories, 2) verify anchors, 3) insert \texttt{<ref\_bbox>} in context. }
    \label{fig:pipeline_instance}
    % \vspace{-1.0em}
\end{figure*}
\subsubsection{PF-GRPO}
\label{subsubsec:pa_grpo}

After ACR-SFT, we further optimize the model with PF-GRPO. 
In this stage, both the policy model $\pi_{\theta}$ and the reference model $\pi_{\mathrm{ref}}$ are initialized from the Stage-1 ACR-SFT checkpoint. 
Unlike the supervised objective in Stage 1, PF-GRPO updates the policy according to the relative quality of a group of sampled reasoning trajectories. 
This optimization incorporates RISS for reasoning-informative sample selection and PAR for process-aware trajectory evaluation.

In the training step $n$, we first construct a candidate pool $\mathcal{C}_n$ by oversampling $k\times m$ samples from the training set:
\begin{equation}
\mathcal{C}_n \sim \mathrm{Sample}_{k m}(\mathcal{D}_{\mathrm{acr}}).
\end{equation}
RISS then estimates the expected PAR of each candidate and selects the samples whose predicted values are closest to the target threshold $\tau_{\mathrm{RISS}}$:
\begin{equation}
\mathcal{B}_n
=
\arg\min_{\mathcal{S}\subset\mathcal{C}_n, |\mathcal{S}|=m}
\sum_{x\in\mathcal{S}}
\left|
V_{\phi}(x)-\tau_{\mathrm{RISS}}
\right|,
\end{equation}
where $V_{\phi}$ is the RISS value model, $m$ is the selected batch size.
This sampling strategy allocates more budget to samples that are more informative for improving reasoning quality.

For each selected sample $x=(I,Q)\in\mathcal{B}_n$, we sample a group of $G$ candidate trajectories $\{o_i\}_{i=1}^{G}$ from the old policy $\pi_{\theta_{\mathrm{old}}}$. 
Each trajectory is evaluated by the total reward:
\begin{equation}
R_i = R_{\mathrm{PAR}}(o_i) + w_{\mathrm{fmt}} I_{\mathrm{fmt}}(o_i),
\end{equation}
where $R_{\mathrm{PAR}}$ is the proposed PAR, and $I_{\mathrm{fmt}} \in \{0,1\}$ indicates whether the generated output follows the structural format, which organizes the reasoning and answer with \texttt{\textless thinking\textgreater}...\texttt{\textless /thinking\textgreater} and \texttt{\textless answer\textgreater}...\texttt{\textless /answer\textgreater} tags, and $w_{\mathrm{fmt}}$ denotes the format reward.
Based on the total rewards within each sampled group, the group-relative advantage is computed as
\begin{equation}
\hat{A}_i
=
\frac{R_i-\bar{R}}
{\sigma(\mathbf{R})},
\qquad
\mathbf{R}=\{R_1,\dots,R_G\},
\end{equation}
where $\bar{R}$ and $\sigma(\mathbf{R})$ denote the mean and standard deviation of the group rewards.

The PF-GRPO objective is defined as
\begin{equation}
\mathcal{L}_{\mathrm{PF\text{-}GRPO}}
=
-\mathbb{E}_{x\sim\mathcal{B}_n}
\left[
\frac{1}{G}
\sum_{i=1}^{G}
\ell_i(\theta;\hat{A}_i)
\right]
+
\lambda D_{\mathrm{KL}}(\pi_{\theta}\|\pi_{\mathrm{ref}}),
\end{equation}
where
\begin{equation}
\begin{aligned}
\ell_i(\theta;\hat{A}_i)
=
\frac{1}{|o_i|}
\sum_{t=1}^{|o_i|}
\min\Big(
& r_{i,t}(\theta)\hat{A}_i, \\
& \operatorname{clip}(r_{i,t}(\theta),1-\epsilon,1+\epsilon)\hat{A}_i
\Big).
\end{aligned}
\end{equation}
and
\begin{equation}
r_{i,t}(\theta)=
\frac{\pi_\theta(o_{i,t}|x,o_{i,<t})}
{\pi_{\theta_{\mathrm{old}}}(o_{i,t}|x,o_{i,<t})}.
\end{equation}

Here, $\epsilon$ is the clipping threshold, the clipping term stabilizes policy updates, and the KL regularization prevents the optimized policy $\pi_\theta$ from drifting excessively from the reference policy $\pi_{\mathrm{ref}}$. $\lambda$ controls the strength of KL regularization.
With RISS and PAR-based trajectory evaluation, PF-GRPO goes beyond optimizing only the final grounding result and further encourages a more faithful reasoning process.

\subsection{Anchor-Centric Reasoning Data Preparation and Verification}
\label{subsec:data_pipeline}

To support ACR-SFT, we construct a structured reasoning data pipeline tailored for RSVG, as illustrated in Fig.~\ref{fig:pipeline_instance}. This pipeline not only produces progressive reasoning trajectories with intermediate reference objects, but also verifies whether the reference anchor is semantically and spatially valid, and inserts its coordinates into the reasoning chain in an context aligned manner. Explicit reference anchors are mainly introduced for complex queries. For simple queries, we do not enforce an explicit reference anchor.

\subsubsection{Structured Reasoning Trajectory Synthesis}
\label{subsubsec:reasoning_synthesis}
To obtain Anchor-Centric reasoning data, we exploit the strong deductive capability of Qwen3VL-Plus to synthesize structured reasoning trajectories conditioned on a remote sensing image $I$, a language query $Q$, and the ground truth $B_{\text{tgt}}$.
Unlike existing works\cite{huang2026rsground},\cite{liu2025towards}, we do not let the model freely explore its own solution path for complex queries. Instead, we require the model to decompose complex descriptions into explicit clues and organize them into a progressive search-region narrowing process. 
It first uses coarse spatial clues to reduce the candidate search region, and then localizes a reference object. To introduce an explicit visual anchor, we require the model to identify the reference object mentioned in the query and predict its bounding box $B_{\text{ref}}^{\text{cand}}$. The model then interprets the relative directional or scale relations described in the query with respect to this anchor, thereby further constraining a more localized search region. Within this localized region, the model identifies several candidate objects and progressively eliminates distractors by incorporating finer-grained clues, such as color, appearance, and geometry, until the target object that is most consistent with all clues is determined. For simple queries, we allow the model to freely explore its own solution path, where it jointly integrates the available clues for localization without introducing unnecessary reasoning.

\subsubsection{Anchor-Centric Verification}
\label{subsubsec:anchor_verification}

Although advanced MLLMs have strong generative capability, directly relying on them to synthesize reasoning trajectories for remote sensing imagery still entails substantial risks of spatial hallucination and semantic mismatch. This issue is particularly severe for complex queries involving reference objects and relational constraints, whose generated reasoning chains are usually more fragile and more prone to errors in reference identification, spatial relation understanding, or scale judgment. To ensure that the reference objects are truly and accurately grounded, we focus on verifying the correctness of such reasoning chains.

Specifically, we first employ Qwen3-Max to parse the text query and extract three types of reference-related information: its absolute location, its relative position or scale, and the corresponding reference noun phrase.

\paragraph{Cross-Modal Semantic Consistency}

To ensure that the generated anchor is semantically accurate, we crop the image patch corresponding to $B_{\text{ref}}^{\text{cand}}$ and employ the pretrained RemoteCLIP~\cite{liu2024remoteclip} model to compute the normalized cosine similarity between this visual patch and the extracted reference noun phrase. A reasoning trajectory is retained only when the similarity score exceeds an empirical threshold $\tau_{\text{sem}}$.

\paragraph{Spatial and Scale Consistency}

Beyond semantic matching, we further verify whether the generated anchor satisfies the extracted geometric clues. For absolute-position clues, we partition the image into a $3 \times 3$ grid and check whether the center of the candidate anchor $B_{\text{ref}}^{\text{cand}}$ falls within the designated region, such as ``top-left.'' For relative spatial or scale clues, such as ``lower-left of'' or ``larger than,'' we construct a set of deterministic geometric rules. By computing the relative center offsets and bounding box area ratios between $B_{\text{ref}}^{\text{cand}}$ and the ground truth $B_{\text{tgt}}$, we perform strict geometric consistency verification. 
To maximize data yield while maintaining strict quality control, we further introduce a retry mechanism. If a candidate trajectory fails either consistency check and is filtered out, the model is prompted to generate a new candidate trajectory. Up to three attempts are allowed for each query. Samples that fail all three attempts are permanently discarded, which helps preserve a clean supervision signal for ACR-SFT. 
\subsubsection{Context-Aligned Anchor Insertion}
\label{subsubsec:anchor_insertion}

During the initial reasoning trajectory generation stage, reference coordinates tend to appear near the end of the reasoning chain. This phenomenon is closely related to the generation patterns of current large models~\cite{bai2025qwen3,hong2025glm}. In pretraining and downstream spatial-output tasks such as object detection and referring expression comprehension, coordinates are typically modeled as part of the final answer and are therefore generated near the end of the output sequence. 
However, for autoregressive models, reference coordinates appended only at the end of the reasoning text cannot serve as an effective visual prior for subsequent reasoning steps; instead, they become a post-hoc supplement to an already completed reasoning process.

To address this issue, we introduce a Context-Aligned Insertion mechanism driven by Qwen3-Max. The verified reference coordinate $B_{\text{ref}}^{*}$ is wrapped with the special token \texttt{<ref\_bbox>}. The model then parses the filtered reasoning chain and inserts this coordinate sequence immediately after the position where the reference object is first semantically identified in the reasoning text. 
This context-aligned insertion strategy ensures that the model explicitly predicts and ``sees'' the anchor before generating the later reasoning steps that depend on it. As a result, the anchor can genuinely function as a prior for subsequent local search and relational reasoning.

After anchor verification and context-aligned insertion, we further randomly sample $10\%$ of the retained trajectories for manual inspection and correct residual issues. After this final check and potential data leakage removal, we obtain the final Anchor-Centric Reasoning Data for ACR-SFT.                                            

\subsection{Process-Aware Reward}
\label{subsec:par}

Existing uses of GRPO in visual grounding typically rely on sparse rewards that only evaluate the final grounding outcome. However, such rewards provide insufficient constraints on the intermediate grounding process, because the model may occasionally localize the correct target without following a correct reasoning path. This issue manifests differently across query types. For simple queries, the model may introduce unnecessary intermediate steps. For complex relational queries, it may bypass the intended reference-driven deduction path. 
To better align PF-GRPO with the desired grounding procedure, we introduce PAR, which adaptively evaluates whether a generated trajectory follows an appropriate grounding process under different query types.

To provide a unified formulation across different query types, we decompose PAR into three components: 
a target term $R_{\mathrm{tgt}}$, which rewards final target object localization; 
a coupling term $R_{\mathrm{cpl}}$, which encourages coherent reasoning between reference grounding and target localization; 
and a penalty term $R_{\mathrm{pen}}$, which penalizes reference predictions that are inconsistent with the query structure.

Accordingly, PAR is defined as
\begin{equation}
R_{\mathrm{PAR}}
=
R_{\mathrm{tgt}}
+
R_{\mathrm{cpl}}
-
R_{\mathrm{pen}},
\end{equation}

where the three components are instantiated differently depending on whether the input query requires a reference.

\begin{figure}[t]
\centering
\includegraphics[width=\columnwidth,trim=0 50 0 0, clip]{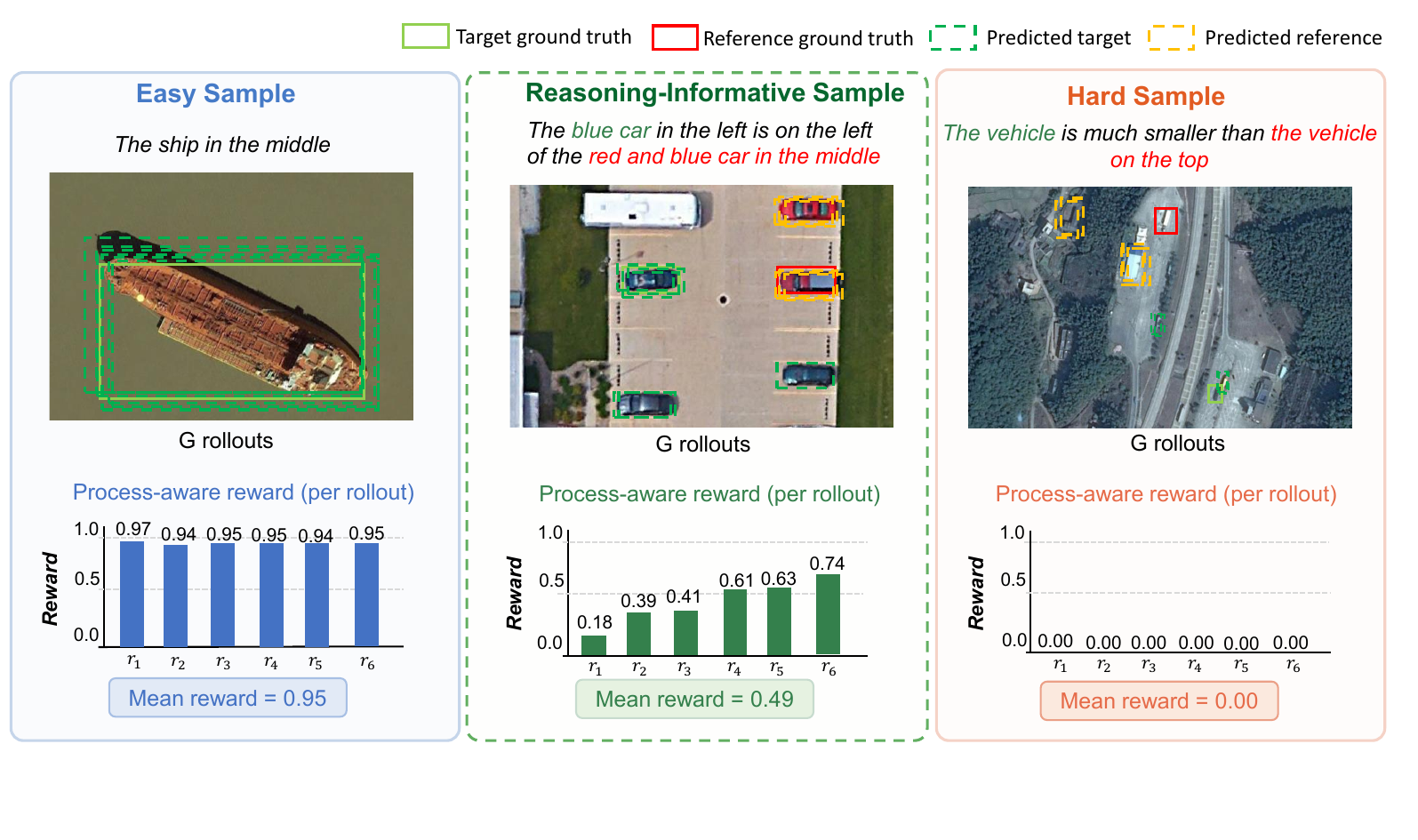}
\caption{Illustration of RISS using observed PAR responses at PF-GRPO step 1. Samples with non-saturated mean PAR are prioritized for more informative process optimization.}
\label{fig:difficulty}
% \vspace{-1.0em}
\end{figure}

\paragraph{Direct Grounding for Simple Queries}

For simple queries that can be resolved by distinctive visual clues alone, the desired grounding process should remain concise and should not introduce unnecessary intermediate references. Accordingly, for simple queries, the target term is defined as the IoU reward for final target object localization, the coupling term is set to zero, and the penalty term penalizes unnecessary reference predictions:
\begin{equation}
R_{\mathrm{tgt}} = \mathrm{IoU}_{\mathrm{tgt}}, \qquad
R_{\mathrm{cpl}} = 0, \qquad
R_{\mathrm{pen}} = \lambda_{\mathrm{hal}} \cdot N_{\mathrm{ref}},
\end{equation}

and the resulting PAR for simple queries becomes
\begin{equation}
R_{\mathrm{PAR}}^{\mathrm{simple}}
=
\mathrm{IoU}_{\mathrm{tgt}}
-
\lambda_{\mathrm{hal}} \cdot N_{\mathrm{ref}}.
\end{equation}

Here, $\lambda_{\mathrm{hal}}$ controls the penalty strength, and $N_{\mathrm{ref}}$ denotes the number of unnecessary reference predictions.

\paragraph{Relational Grounding for Complex Queries}

For complex relational queries, a reliable grounding process requires the model to first ground the reference object, then interpret the position or scale relation, and finally localize the target according to this relation. Considering that the reference coordinate serves as a visual anchor that connects relational reasoning with final target localization, it directly reflects whether the reference object is correctly grounded. Moreover, when both the reference object and the final target are correctly localized, the consistency between them also provides direct evidence for correct relational reasoning. Therefore, we design the reward as follows:
\begin{equation}
R_{\mathrm{tgt}}
=
\alpha_{\mathrm{tgt}} \cdot \mathrm{IoU}_{\mathrm{tgt}},
\end{equation}

\begin{equation}
R_{\mathrm{cpl}}
=
\begin{cases}
\alpha_{\mathrm{ref}} \cdot \mathrm{IoU}_{\mathrm{ref}}
+
\beta \cdot \mathbb{I}_{\mathrm{bonus}},
& N_{\mathrm{ref}} = 1, \\[2pt]
0,
& \text{otherwise},
\end{cases}
\end{equation}

\begin{equation}
R_{\mathrm{pen}}
=
\lambda_{\mathrm{multi}} \cdot |N_{\mathrm{ref}} - 1|,
 \\[2pt]
\end{equation}

The resulting PAR for relational queries is
\begin{equation}
R_{\mathrm{PAR}}^{\mathrm{rel}}
=
R_{\mathrm{tgt}}
+
R_{\mathrm{cpl}}
-
R_{\mathrm{pen}}.
\end{equation}

Here, $\alpha_{\mathrm{tgt}}$ and $\alpha_{\mathrm{ref}}$ balance the contributions of final target localization and intermediate reference grounding. When the model produces the desired trajectory with exactly one reference prediction, the coupling term rewards the quality of reference grounding and activates an additional bonus only when both the reference and target predictions exceed predefined thresholds 0.5. If the model omits the required reference object or generates multiple redundant references, the penalty term is activated according to the deviation of $N_{\mathrm{ref}}$ from the expected reference count. 

In this way, PAR does not merely encourage correct target coordinates. It also encourages the model to follow a grounding process consistent with the relational structure of the query.

\begin{algorithm}[!t]
\small
\caption{RISS in PF-GRPO}
\label{alg:rids}
% \vspace{-0.3em}
\begin{algorithmic}[1]
\REQUIRE Dataset $\mathcal{D}$, selected query number $m$, generations per query $G$, target threshold $\tau_{\mathrm{RISS}}$, oversampling factor $k$
\STATE \textbf{Initialize:} policy model $\pi_{\theta_0}$, value model $V_{\phi_0}$
\FOR{$n=0$ to $N-1$}
    \STATE Sample an overscaled candidate pool:
    
    $\mathcal{D}_{km}=\{(I^i,Q^i)\}_{i=1}^{k\times m}\sim\mathcal{D}$
    \STATE Predict expected PAR for each candidate:
    $v^i = V_{\phi_n}(I^i,Q^i)$
    \STATE Select $m$ samples $\mathcal{D}_m \subset \mathcal{D}_{km}$ that minimize $|v^i-\tau_{\mathrm{RISS}}|$
    \STATE Generate $G$ trajectories for each selected query:
    
    $\{T^{i,j}\}_{j=1}^{G}\sim\pi_{\theta_n}(\cdot\mid I^i,Q^i)$
    \STATE Compute $R_{\mathrm{PAR}}^{i,j}$ for all generated trajectories
    \STATE Update policy $\pi_{\theta_n}\rightarrow\pi_{\theta_{n+1}}$ using the PF-GRPO objective
    \STATE Update value model $V_{\phi_n}\rightarrow V_{\phi_{n+1}}$ by minimizing
    
    $\mathcal{L}_{\mathrm{value}}
    =
    \frac{1}{m}\sum_{i=1}^{m}
    \left(
    V_{\phi_n}(I^{i},Q^{i})
    -
    \frac{1}{G}\sum_{j=1}^{G}R_{\mathrm{PAR}}^{i,j}
    \right)^2$
\ENDFOR
\end{algorithmic}
% \vspace{-0.6em}
\end{algorithm}
\subsection{Reasoning-Informative Sample Selector}
\label{subsec:rids}
PAR measures the quality of the reasoning behavior performed by the current model on a given sample, and is later used as the reward signal for policy updates.
As shown in Fig.~\ref{fig:difficulty}, we observe that, at a given training step, samples with different difficulty levels induce markedly different PAR responses over sampled reasoning trajectories. Since the raw PAR is not necessarily bounded within $[0,1]$, we normalize it before RISS selection. Unless otherwise specified, PAR values in this subsection refer to the normalized PAR.

For samples whose required reasoning has been largely mastered by the current model, most rollouts tend to receive consistently high PAR, so further optimization on these samples provides limited further benefit. For samples that are still too difficult at the current training step, most rollouts tend to receive consistently low PAR, making it difficult to provide effective positive feedback for PF-GRPO. 
In contrast, samples with moderate mean PAR are more informative for the current model. Such samples are neither stably solved nor consistently failed, suggesting that the model has not yet fully mastered the required reference grounding, relation understanding, or target localization capability, but can already receive useful positive feedback from them.
Therefore, the mean PAR provides a practical proxy for selecting samples that are more suitable for reasoning optimization at the current training step.

To this end, we introduce RISS to estimate the expected PAR response of candidate samples during data sampling and prioritize samples with moderate predicted PAR values, thereby improving the effectiveness of PF-GRPO for progressive reasoning optimization.
Specifically, RISS is driven by an online value model built on a Qwen3VL-2B backbone with an additional lightweight regression head.
During each RL iteration, as outlined in Algorithm~\ref{alg:rids}, we first oversample a candidate pool of size $k \times m$ from the training set.
The value model then estimates the expected PAR response of all candidates, and we select the $m$ samples whose predicted values are closest to a target threshold $\tau_{\mathrm{RISS}}$.
We set $\tau_{\mathrm{RISS}}$ to 0.4, and the sensitivity analysis in Section~\ref{subsec:sensitivity} further supports this choice.
Subsequently, the model generates $G$ rollout trajectories only for the selected samples, and PF-GRPO updates are performed based on PAR.
Meanwhile, the value model is updated online by minimizing the mean squared error between its prediction and the observed mean PAR over the sampled trajectories.

It is worth noting that RISS does not permanently discard hard samples.
As the model becomes stronger, samples that were previously too difficult may move from the low-PAR saturated region to the non-saturated region, and can therefore be prioritized in later training stages.

\begin{table}[t!]
\centering
\caption{Statistics of the original RSVG datasets and the Anchor-Centric Reasoning Data used for ACR-SFT. The reduction in training set size comes from strict trajectory verification and leakage removal during data construction.}
\label{tab:dataset_stats}
\resizebox{\columnwidth}{!}{
\begin{tabular}{l c c c c}
\toprule
\multirow{2}{*}{\textbf{Dataset}} & \multicolumn{2}{c}{\textbf{Original Dataset}} & \multicolumn{2}{c}{\textbf{Anchor-Centric Reasoning Data}} \\
\cmidrule(lr){2-3} \cmidrule(lr){4-5}
& Train (Pairs) & Test (Pairs) & Train (Pairs) & Test (Pairs) \\
\midrule
DIOR-RSVG & 26,991 & 7,500 & \textbf{25,012} & \textbf{7,500} \\
OPT-RSVG & 19,580 & 24,477 & \textbf{18,875} & \textbf{24,477} \\
VRS-Bench & 36,313 & 16,159 & \textbf{36,106} & \textbf{16,159} \\
\bottomrule
\end{tabular}
}
\end{table}
\section{Experiments}
In this section, we first describe the datasets, evaluation metrics, and implementation details in Sections~\ref{subsec:datasets}, \ref{subsec:metrics}, and~\ref{subsec:implementation}. We then report the main comparison results in Section~\ref{subsec:sota}. Finally, we present ablation studies, hyperparameter sensitivity analysis, and an analysis of reference anchor quality in Sections~\ref{subsec:ablation}, \ref{subsec:sensitivity}, and~\ref{subsec:anchor_quality}, respectively.
\subsection{Datasets}
\label{subsec:datasets}

We evaluate GeoSearcher on three widely used benchmarks: DIOR-RSVG, OPT-RSVG, and VRS-Bench. These datasets covers base grounding, complex scene understanding, and fine-grained reasoning.

\textbf{DIOR-RSVG}\cite{zhan2023rsvg} is a foundational benchmark built on the DIOR dataset. With a fixed resolution of $800 \times 800$, it offers a relatively standardized setting for assessing basic grounding performance. Its queries involve both object attributes and spatial expressions. Following the standard split, we use 26,991 samples for training, 3,829 for validation, and 7,500 for testing.

\textbf{OPT-RSVG}\cite{li2024language} is a more challenging benchmark for evaluating robustness under diverse observation conditions and image scales. Compared with DIOR-RSVG, it contains more complex target scales, more similar objects within a scene, more varied viewpoints and color appearance, and noisier backgrounds. These factors make target recognition and spatial disambiguation more difficult. We follow the official protocol and use 19,580 training samples, 4,895 validation samples, and 24,477 test samples.

\textbf{VRS-Bench}\cite{li2024vrsbench} is designed for fine-grained reasoning. It uses standardized $512 \times 512$ image patches and high-quality human-verified annotations, which provide reliable evaluation signals. Its queries place stronger emphasis on precise alignment among object attributes, spatial relations, and scene context. We adopt the standard split with 36,313 samples for training and 16,159 for testing.

Table~\ref{tab:dataset_stats} reports the statistics of the original RSVG datasets and the final Anchor-Centric Reasoning Data used for ACR-SFT. For ACR-SFT, we construct Anchor-Centric Reasoning Data from the official training partitions of each benchmark rather than directly using the original training splits. 
Before supervised fine-tuning, all synthesized trajectories are processed by verification, manual inspection, and leakage removal. 
In particular, some RSVG benchmarks contain cases where the same remote sensing image appears in both the training and test partitions with different language queries. 
Since ACR-SFT introduces explicit reference coordinates into the reasoning trajectories, such image-level overlap may expose the model to object locations from test images. 
To avoid this potential leakage, we remove all training samples whose image paths overlap with the test set. 
\subsection{Evaluation Metrics}
\label{subsec:metrics}

Following prior works, we adopt Pr@0.5, Pr@0.7, and mIoU as the target-level grounding metrics.
Specifically, Pr@0.5 and Pr@0.7 denote the percentages of samples whose IoU between the predicted target bounding boxes and the ground truth target bounding boxes exceeds 0.5 and 0.7, respectively.

The mIoU is defined as
\begin{equation}
\text{mIoU}=\frac{1}{M}\sum_{t=1}^{M}\frac{I_t}{U_t},
\end{equation}
where $M$ is the number of samples, and $I_t$ and $U_t$ denote the intersection and union areas between the predicted and ground truth target bounding boxes of the $t$-th sample, respectively.

To evaluate process correctness on relational queries, we further introduce two process-level metrics: Faithful@0.5 and Shortcut@0.5.
These metrics are computed only on the relational-query subset $\mathcal{R}$ and are applicable only to models that explicitly generate reference object bounding boxes.

Let
\begin{equation}
\begin{aligned}
z_i^{\mathrm{ref}}
&=\mathbf{1}\!\left[\mathrm{IoU}(B^{\mathrm{pred}}_{\mathrm{ref}},B^{\mathrm{gt}}_{\mathrm{ref}})>0.5\right],\\
z_i^{\mathrm{tgt}}
&=\mathbf{1}\!\left[\mathrm{IoU}(B^{\mathrm{pred}}_{\mathrm{tgt}},B^{\mathrm{gt}}_{\mathrm{tgt}})>0.5\right].
\end{aligned}
\end{equation}

Then Faithful@0.5 is defined as
\begin{equation}
\text{Faithful@0.5}
=
\frac{1}{|\mathcal{R}|}\sum_{i\in\mathcal{R}} z_i^{\mathrm{ref}} z_i^{\mathrm{tgt}}.
\end{equation}

Shortcut@0.5 is defined as
\begin{equation}
\text{Shortcut@0.5}
=
\frac{1}{|\mathcal{R}|}\sum_{i\in\mathcal{R}} (1-z_i^{\mathrm{ref}})\, z_i^{\mathrm{tgt}}.
\end{equation}

To compute process-level metrics that require reference object localization, we additionally construct reference anchor annotations for the test sets using the same pipeline as the training data construction.
All bounding boxes are manually corrected and verified to ensure accurate reference grounding. 
The resulting human-verified reference bounding boxes are used for evaluation, including Faithful@0.5, Shortcut@0.5, and the reference anchor quality analysis.

\begin{figure}[t]
\centering
\includegraphics[
    width=\columnwidth,
    trim=20 350 20 0,
    clip
]{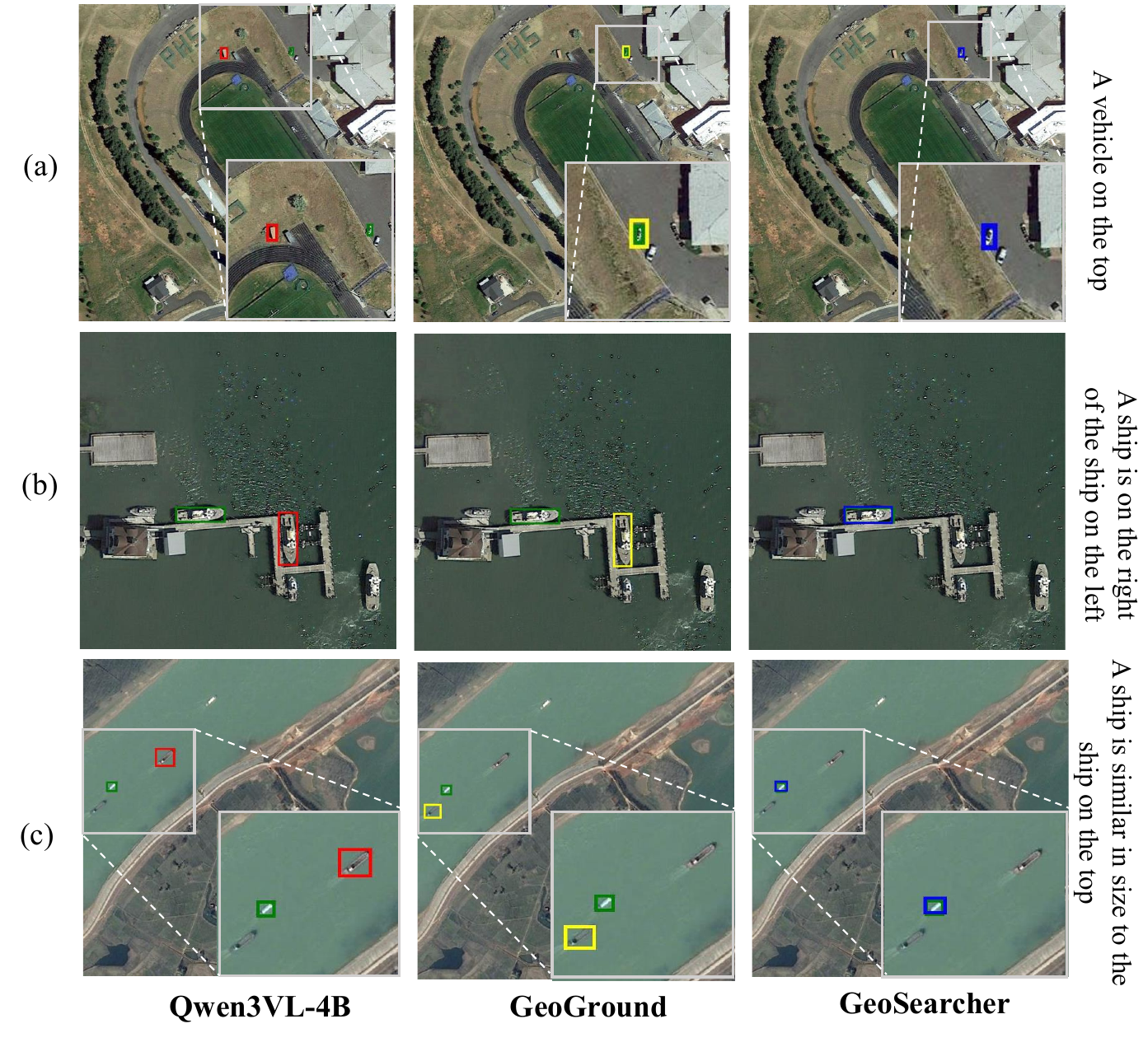}
\caption{Qualitative comparison with representative baselines on different query types. From left to right are Qwen3VL-4B, GeoGround, and our GeoSearcher. Green boxes denote ground truth, while colored boxes denote predictions from different methods.}
\label{fig:qualitative_comp}
% \vspace{-1.2em}
\end{figure}
\subsection{Implementation Details}
\label{subsec:implementation}

GeoSearcher uses Qwen3VL-4B as the base model. The overall training pipeline is implemented with LLaMA-Factory\cite{zheng2024llamafactory} for supervised fine-tuning and Verl\cite{sheng2025hybridflow} for reinforcement learning.
In Stage 1, we perform ACR-SFT on the Anchor-Centric Reasoning data for 3 epochs with a learning rate of $1\times10^{-5}$ and a global batch size of 64. In the semantic consistency check, the RemoteCLIP similarity threshold $\tau_{\text{sem}}$ is set to 0.85. We adopt DeepSpeed ZeRO-3\cite{rasley2020deepspeed} for memory-efficient training. The cutoff length is set to 2048, the learning rate scheduler follows a cosine schedule, and the warmup ratio is set to 0.1. 
In Stage 2, we conduct PF-GRPO for 200 steps with a learning rate of $1\times10^{-6}$ and a global batch size of 128. During policy optimization, we sample $G=6$ trajectories for each query. For PAR, we set $\lambda_{\mathrm{hal}}=0.2$ and $\lambda_{\mathrm{multi}}=0.2$. 
The target and reference rewards in the coupling design are weighted by 
$\alpha_{\mathrm{tgt}}=0.7$ and $\alpha_{\mathrm{ref}}=0.3$, respectively, as determined by hyperparameter tuning. 
The bonus coefficient is set to $\beta=0.2$. In addition, the format reward weight is set to $w_{\mathrm{fmt}}=0.2$.
For the RISS, we freeze the Qwen3VL-2B backbone parameters and train only lightweight LoRA adapters together with the regression head. In our experiments, the pool expansion factor is set to $k=2$.
All experiments are conducted on 2 NVIDIA A100 GPUs. 

\begin{table*}[t]
\centering
\caption{Quantitative Comparison of the GeoSearcher Against State-of-the-Art Models on Three Benchmarks: DIOR-RSVG, OPT-RSVG, and VRS-Bench. The best overall results are highlighted in \textbf{bold}, and the second results are \underline{underlined}.}
\label{tab:main_results_comprehensive}
\resizebox{\textwidth}{!}{
\begin{tabular}{l ccc ccc ccc}
\toprule
\multirow{2}{*}{\textbf{Method}} & \multicolumn{3}{c}{\textbf{DIOR-RSVG}} & \multicolumn{3}{c}{\textbf{OPT-RSVG}} & \multicolumn{3}{c}{\textbf{VRS-Bench}} \\
\cmidrule(lr){2-4} \cmidrule(lr){5-7} \cmidrule(lr){8-10}
& Pr@0.5 & Pr@0.7 & mIoU & Pr@0.5 & Pr@0.7 & mIoU & Pr@0.5 & Pr@0.7 & mIoU \\
\midrule
\multicolumn{10}{l}{\textit{General MLLMs}} \\
ChatGPT-5\cite{singh2025openai} & 27.2 & 4.2 & 29.6 & 20.6 & 10.1 & 21.2 & 14.9 & 3.4 & 23.6 \\
MiniGPT-v2-7B\cite{chen2023minigptv2largelanguagemodel} & 30.3 & 11.5 & 30.1 & 21.4 & 11.2 & 23.2 & 35.8 & 16.8 & 34.7 \\
Qwen2.5VL-3B\cite{bai2025qwen25vltechnicalreport} & 33.7 & 20.4 & 33.5 & 24.6 & 15.2 & 25.3 & 37.1 & 20.3 & 34.9 \\
GLM-4.1V-9B-Thinking\cite{hong2025glm} & 47.7 & 33.6 & 45.3 & 43.6 & 29.1 & 43.3 & 46.6 & 24.9 & 42.7 \\
Qwen3VL-4B\cite{bai2025qwen3} & 50.7 & 37.4 & 47.4 & 43.2 & 29.3 & 41.6 & 56.1 & 36.4 & 49.6 \\
Qwen3.5-4B & 53.1 & 39.4 & 49.0 & 43.9 & 30.1 & 43.1 & 47.4 & 23.6 & 43.6 \\
InternVL3.5-8B\cite{wang2025internvl3} & 64.9 & 50.5 & 56.9 & 46.3 & 32.6 & 44.2 & 54.9 & 29.3 & 48.8 \\
\midrule
\multicolumn{10}{l}{\textit{RS-Specific MLLMs}} \\
Geochat\cite{kuckreja2024geochat} & 19.7 & - & - & 13.4 & 2.9 & 20.8 & 57.4 & 22.6 & - \\
SkysenseGPT\cite{luo2024skysensegpt} & 22.9 & 5.5 & 27.8 & 5.5 & 1.1 & 13.9 & 11.0 & 2.2 & 21.3 \\
EarthDial\cite{soni2025earthdial} & 46.1 & 30.2 & 39.5 & 29.6 & 15.8 & 30.1 & 14.4 & 7.8 & 13.0 \\
VHM\cite{pang2025vhm} & 56.2 & 36.0 & 50.3 & 31.8 & 12.8 & 33.2 & 23.5 & 4.4 & 28.9 \\
EarthGPT\cite{zhang2024earthgpt} & 76.7 & \underline{66.5} & \underline{69.3} & - & - & - & - & - & - \\
GeoGround\cite{zhou2024geoground} & \underline{77.7} & - & - & \underline{53.3} & \underline{39.6} & \underline{48.2} & \underline{66.0} & - & - \\
\midrule
\multicolumn{10}{l}{\textit{RS Reasoning MLLMs}} \\
Geo-R1\cite{zhang2025geo} & 40.6 & - & - & 34.2 & 20.6 & 35.3 & 59.5 & 37.1 & - \\
RSGround-R1\cite{huang2026rsground} & 71.8 & 58.7 & 63.4 & - & - & - & 63.7 & \underline{38.3} & \underline{54.7} \\
TinyRS-R1\cite{AyboraKksal2025TinyRSR1CV} & 74.9 & - & - & 38.3 & 25.3 & 36.1 & 15.1 & 6.0 & 18.6 \\
GeoZero\cite{wang2025geozero} & 75.7 & - & - & - & - & - & - & - & - \\
\midrule
\textbf{GeoSearcher} & \textbf{83.2} & \textbf{72.4} & \textbf{73.5} & \textbf{81.2} & \textbf{65.5} & \textbf{69.4} & \textbf{80.3} & \textbf{64.4} & \textbf{69.2} \\
\bottomrule
\end{tabular}
}
\end{table*}

\subsection{Comparison with State-of-the-Art Methods}
\label{subsec:sota}

Table~\ref{tab:main_results_comprehensive} presents a quantitative comparison between GeoSearcher and a broad range of state-of-the-art baselines, including general MLLMs, RS-specific MLLMs, and recent RS reasoning models. 
For general MLLMs, we re-evaluate the models under a unified protocol using $1024 \times 1024$ inputs. For RS-specific MLLMs and RS reasoning models, we use the results reported in their original papers whenever available; otherwise, we re-test them under the same $1024 \times 1024$ protocol.

On DIOR-RSVG, GeoSearcher reaches 83.2\% at Pr@0.5, surpassing the strongest previous baseline, GeoGround, by 5.5\%. The gain becomes much larger on OPT-RSVG, where scenes are larger and the descriptions are more complex, with GeoSearcher improving Pr@0.5 by 27.9\%. These results indicate that GeoSearcher is particularly effective when localization must be performed under a larger visual search space, where multiple clues need to be integrated to progressively reduce ambiguity.
GeoSearcher also delivers the best overall performance on VRS-Bench, where the native image resolution is relatively modest but the language descriptions are more complex, exceeding GeoGround by a clear margin of 14.3\%. This suggests that the advantage of GeoSearcher is not limited to search-space reduction in high-resolution imagery, but also stems from its stronger ability to integrate complex semantic clues during grounding.

\begin{figure}[t]
\centering
\includegraphics[width=\columnwidth,trim=0 0 10 0,
    clip]{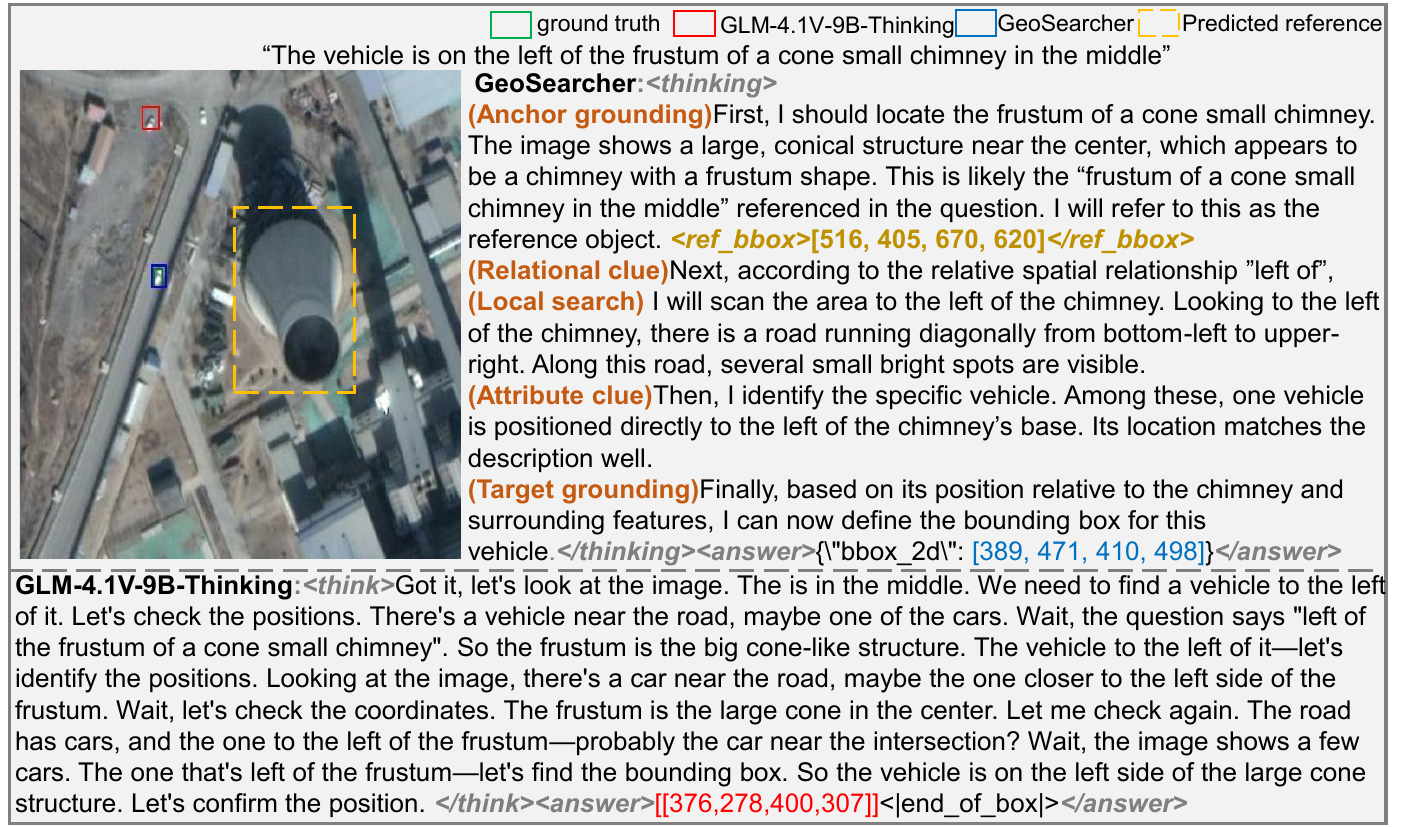}
\caption{Case study of reasoning behavior. Compared with a strong reasoning baseline.}
\label{fig:qualitative}
% \vspace{-0.8em}
\end{figure}

These results are reinforced by the qualitative analysis
presented in Fig.~\ref{fig:qualitative_comp}. GeoSearcher produces predictions that are consistently closer to the ground truth across three representative query types, while different baselines exhibit distinct failure patterns.
In row a, because the target is extremely small, Qwen3VL-4B fails to place its prediction on an actual vehicle. In comparison, both GeoGround and GeoSearcher correctly localize the object.
A more challenging case is observed in row b, where the query depends on a relative relation to a small and easily overlooked reference object. Qwen3VL-4B and GeoGround both predict incorrect bounding boxes. Their errors may stem from two possible reasons: they may fail to identify the "the ship on the left", or they may identify it but fail to correctly resolve the spatial relation. In particular, their predicted regions are closer to the lower-right of the reference object rather than strictly to its right, suggesting that the relative spatial constraint is not reliably resolved. In contrast, GeoSearcher correctly grounds the "the ship on the left" as an anchor, and then searches the corresponding right-side region to localize the target accurately.
The last row involves a subtle scale comparison, requiring the model to find ``a ship similar in size to the ship on the top.'' Neither Qwen3VL-4B nor GeoGround accurately localizes "the small ship at the top" as the scale reference, and their predicted targets are therefore slightly larger than the top ship. GeoSearcher, however, still succeeds by making more accurate use of the reference object and the relational clue.

To further illustrate the reasoning process of GeoSearcher, Fig.~\ref{fig:qualitative} provides a representative example. Although GLM-4.1V-9B-Thinking generates a plausible reasoning trace, its process is relatively unstructured and weakly grounded. It mentions the reference object clue, but does not ground it as a critical clue. The subsequent localization process is not tightly centered on this reference object, leading to a final prediction that does not strictly satisfy the ``on its left'' condition.
In contrast, GeoSearcher follows a structured anchor-guided progressive reasoning process. It first grounds the frustum-shaped chimney as the reference anchor, then restricts the search to the local region on its left, and finally identifies the target vehicle by jointly using the relational clue ``left of'' and the attribute clues. This comparison suggests that the improvement of GeoSearcher comes not merely from producing longer reasoning text, but from explicitly grounding an intermediate reference object and using it to guide local target search.

\subsection{Ablation Studies}
\label{subsec:ablation}

In order to thoroughly evaluate the contribution of our proposed GeoSearcher, we conducted a comprehensive series of ablation experiments. 
Specifically, we investigate the effectiveness of the two-stage post-training strategy, the Context-Aligned Anchor Insertion strategies during ACR-SFT, the individual components of the PAR, and the gains from the RISS. 
To ensure computational efficiency, all ablation experiments in this section are conducted at a resolution of $448 \times 448$.

\begin{table}[t!]
\centering
\caption{Ablation study on the two-stage strategy of GeoSearcher on DIOR-RSVG. SFT-only variants are trained for 4 epochs. The two-stage variants use 3 epochs of supervised initialization followed by 200 RL steps.}
\label{tab:core_ablation}
\resizebox{\linewidth}{!}{
\begin{tabular}{@{}l c c c c@{}}
\toprule
\textbf{Model Variant} & \textbf{Data Format} & \textbf{Pr@0.5} & \textbf{Pr@0.7} & \textbf{mIoU} \\
\midrule
SFT Only & Standard & 80.02 & 68.72 & 70.44 \\
ACR-SFT Only & Anchor-Centric Reasoning & 79.64 & 67.80 & 69.62 \\
Basic GRPO Only & Reasoning & 74.21 & 65.15 & 66.20 \\
\midrule
SFT + Basic GRPO & Standard & 80.23 & 69.12 & 70.52 \\
\textbf{ACR-SFT + PF-GRPO} & \textbf{Anchor-Centric Reasoning} & \textbf{81.52} & \textbf{70.12} & \textbf{71.53} \\
\bottomrule
\end{tabular}
}
\end{table}

\subsubsection{Effectiveness of the Two-Stage Post-Training Strategy}
\label{subsubsec:agppt_effectiveness}

Table~\ref{tab:core_ablation} evaluates the effectiveness of the two-stage post-training strategy under a comparable computational budget. Directly applying GRPO to the base model performs poorly, yielding only 74.21\% on Pr@0.5. This suggests that reinforcement learning alone is insufficient for stable grounding ability when the model has not yet acquired a structured reasoning prior.
In contrast, under pure supervised fine-tuning, standard SFT performs slightly better than ACR-SFT. This indicates that anchor-centric reasoning trajectories do not necessarily translate into immediate gains in final localization under supervised learning alone, since the model may learn to imitate intermediate reasoning patterns without yet fully converting them into more accurate target grounding.

Once GRPO is introduced on top of SFT initialization, performance improves consistently. The two-stage variant based on standard SFT reaches 80.23\%, 69.12\%, and 70.52\% on Pr@0.5, Pr@0.7, and mIoU, while the full ACR-SFT + PF-GRPO variant further improves the results to 81.52\%, 70.12\%, and 71.53\%, respectively. This shows that ACR-SFT provides a more suitable starting point for subsequent policy optimization, even though its standalone supervised performance is slightly lower than standard SFT.
Overall, these results validate the effectiveness of the two-stage design in GeoSearcher. ACR-SFT and PF-GRPO play complementary roles: ACR-SFT injects the progressive reasoning prior, while PF-GRPO further optimizes the model to exploit this prior for more accurate and faithful grounding.

\begin{table*}[t]
\centering
\caption{Evaluation of explicit reference coordinate insertion strategies on DIOR-RSVG. ``Simple-Queries'' denotes samples that do not require a reference object, while ``Complex-Queries'' denotes samples that require reference-based reasoning.}
\label{tab:anchor_insertion_unified}
\resizebox{\textwidth}{!}{
\begin{tabular}{@{}l ccc ccc cc@{}}
\toprule
\multirow{2}{*}{\textbf{Strategy}} 
& \multicolumn{3}{c}{\textbf{Simple-Queries}}
& \multicolumn{3}{c}{\textbf{Complex-Queries}}
& \multicolumn{2}{c}{\textbf{Process-level Metrics}} \\
\cmidrule(lr){2-4} \cmidrule(lr){5-7} \cmidrule(lr){8-9}
& \textbf{Pr@0.5} & \textbf{Pr@0.7} & \textbf{mIoU}
& \textbf{Pr@0.5} & \textbf{Pr@0.7} & \textbf{mIoU}
& \textbf{Faithful@0.5 $\uparrow$} & \textbf{Shortcut@0.5 $\downarrow$} \\
\midrule
No explicit reference coordinate & 84.04 & 71.12 & 73.34 & 62.80 & 52.08 & 55.68 & -- & -- \\
Suffix-Appended (end of reasoning) & 84.05 & 72.07 & 73.41 & 64.39 & 53.05 & 56.87 & 43.96 & 20.43 \\
Context-Aligned (after semantic mention) & \textbf{84.30} & \textbf{72.13} & \textbf{73.88} & \textbf{64.82} & \textbf{54.01} & \textbf{57.14} & \textbf{46.38} & \textbf{18.44} \\
\bottomrule
\end{tabular}
}
\end{table*}

\begin{table*}[htbp]
\centering
\caption{Ablation study on the PAR design of GeoSearcher on DIOR-RSVG. All RL variants are initialized from ACR-SFT.}
\label{tab:reward_ablation}
\resizebox{\textwidth}{!}{
\begin{tabular}{@{}l ccc ccc cc@{}}
\toprule
\multirow{2}{*}{\textbf{Stage / Setting}} & \multicolumn{3}{c}{\textbf{Reward Components}} & \multicolumn{3}{c}{\textbf{Outcome Metrics}} & \multicolumn{2}{c}{\textbf{Process Metrics}} \\
\cmidrule(lr){2-4} \cmidrule(lr){5-7} \cmidrule(l){8-9}
& \textbf{Target} & \textbf{Coupling} & \textbf{Penalty} & \textbf{Pr@0.5 (\%)} & \textbf{Pr@0.7 (\%)} & \textbf{mIoU} & \textbf{Faithful@0.5 $\uparrow$} & \textbf{Shortcut@0.5 $\downarrow$} \\
\midrule
ACR-SFT & -- & -- & -- & 79.37 & 67.57 & 69.63 & 46.38 & 18.44 \\
\midrule
Basic GRPO & \cmark & \xmark & \xmark & 80.52 (\(\uparrow\)1.15) & 69.41 (\(\uparrow\)1.84) & 70.75 (\(\uparrow\)1.12) & 51.03 & 15.42 \\
+Coupling & \cmark & \cmark & \xmark & 81.02 (\(\uparrow\)1.65) & 69.61 (\(\uparrow\)2.04) & 71.22 (\(\uparrow\)1.59) & 54.01 & 13.03 \\
+Penalty & \cmark & \xmark & \cmark & 80.71 (\(\uparrow\)1.34) & 69.44 (\(\uparrow\)1.87) & 70.89 (\(\uparrow\)1.26) & 52.35 & 14.53 \\
+PAR & \cmark & \cmark & \cmark & \textbf{81.16} (\(\uparrow\)1.79) & \textbf{69.65} (\(\uparrow\)2.08) & \textbf{71.31} (\(\uparrow\)1.68) & \textbf{54.47} & \textbf{12.58} \\
\bottomrule
\end{tabular}
}
\end{table*}
\subsubsection{Effect of Context-Aligned Anchor Insertion}
\label{subsubsec:insertion_ablation}

Table~\ref{tab:anchor_insertion_unified} compares three strategies for explicit reference coordinate insertion: no explicit reference coordinates, suffix-appended insertion, and the proposed context-aligned insertion. The differences among the three strategies are small on \textit{Simple-Queries}, indicating that explicit reference coordinates are not crucial for queries that do not rely on reference-based reasoning. By contrast, the gains become much more evident on \textit{Complex-Queries}, where context-aligned insertion achieves the best performance. This shows that explicit reference coordinates are most useful when an intermediate anchor is genuinely required, and that their temporal placement within the reasoning chain further affects how effectively they can be used.
The process-level metrics further clarify this effect. Context-aligned insertion yields higher Faithful@0.5 and lower Shortcut@0.5, indicating that its benefit does not merely come from more correct final target bounding boxes, but from more reliably grounding the reference object. This observation is consistent with our design motivation. If reference coordinates are appended only at the end of the reasoning chain, they are more likely to act as a post-hoc supplement rather than a visual prior for subsequent reasoning. In contrast, context-aligned insertion places the coordinates immediately after the semantic mention of the reference object, making the order of reference semantics, explicit grounding, and subsequent relational deduction more causally consistent. As a result, it improves both final performance and process faithfulness.

\subsubsection{Ablation of the PAR}
\label{subsubsec:reward_ablation}
Table~\ref{tab:reward_ablation} reports the ablation results of the proposed PAR design. Starting from the target-only GRPO reward, adding the coupling term improves Pr@0.5 to 81.02\%, while also increasing Faithful@0.5 by 2.98\% and reducing Shortcut@0.5 by 2.39\%. Further introducing the penalty term raises Pr@0.5 to 81.16\%, with Faithful@0.5 and Shortcut@0.5 further improved by 0.46\% and 0.45\%, respectively. This shows that the proposed reward not only improves final grounding performance, but also enhances the faithfulness of the intermediate reasoning process and suppresses shortcut behavior.

\subsubsection{Effect of the RISS}
\label{subsubsec:dds_ablation}

Table~\ref{tab:dds_ablation} evaluates the effect of RISS. 
Introducing RISS brings consistent improvements under both the basic GRPO setting and the full PAR setting, with particularly clear gains in the process-level metrics. 
This indicates that RISS can stably select samples that are more meaningful for reasoning process optimization.
The improvement becomes more evident when RISS is combined with PAR. The reason is that a target-only reward can only indicate whether the final localization has been solved. Since RISS prioritizes samples with moderate reward responses, using only the target reward may overlook samples that still have substantial process-level learning value. For instance, samples with near-one target rewards may still suffer from inaccurate reference grounding or weak relation understanding, while samples with near-zero target rewards may still provide useful intermediate reasoning feedback. In contrast, PAR jointly evaluates reference grounding and final target localization, making process-informative samples more likely to remain in the moderate reward range prioritized by RISS.
These results suggest that RISS and PAR are complementary: PAR provides process-aware reward signals, while RISS focuses optimization on samples where these signals are more useful for learning progressive reasoning.

\begin{figure}[t!]
\centering
\includegraphics[width=1\linewidth]{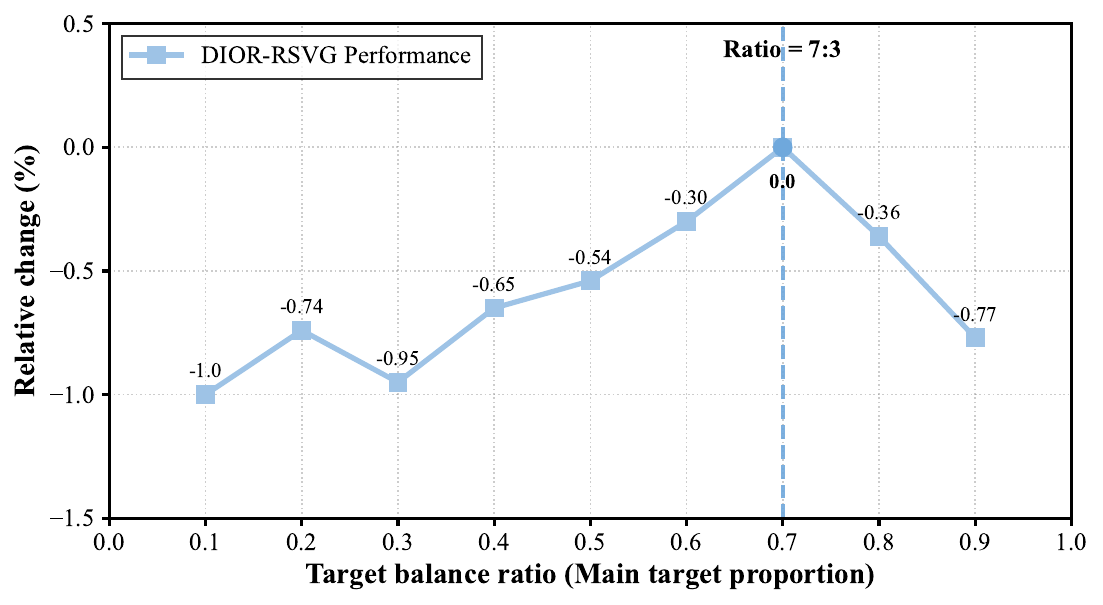}
\caption{Sensitivity analysis of the target-reference reward weight ratio in PAR. The best performance is achieved around $[\alpha_{\mathrm{tgt}}, \alpha_{\mathrm{ref}}]=[0.7,0.3]$.}
\label{fig:alpha_curve}
\end{figure}

\begin{figure}[!t]
\centering
\includegraphics[width=1\linewidth]{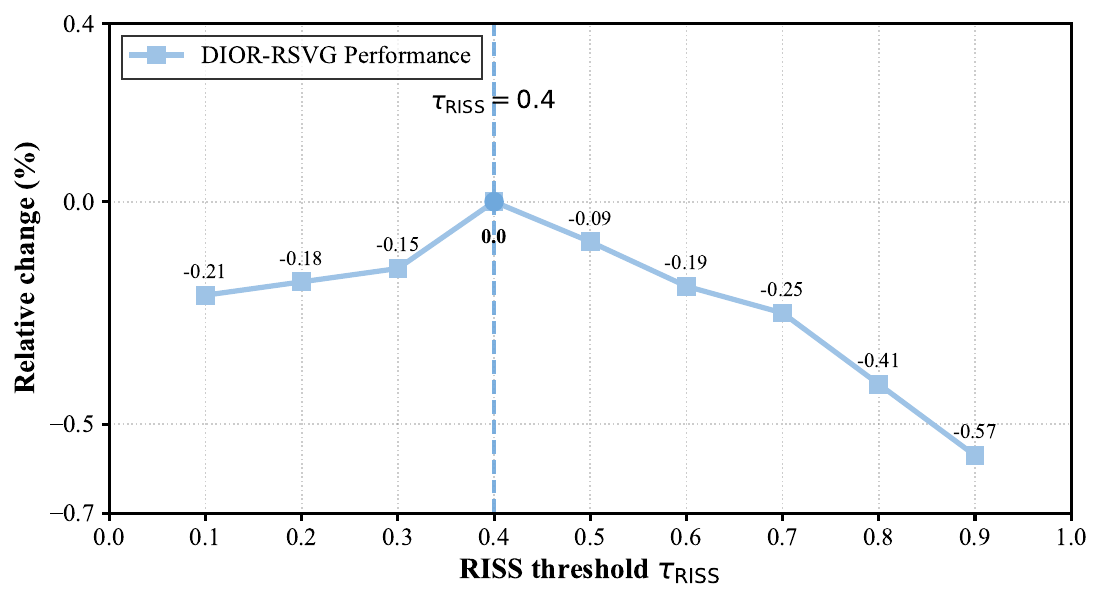}
\caption{Sensitivity analysis of the RISS threshold on DIOR-RSVG.}
\label{fig:rids_threshold}
% \vspace{-0.8em}
\end{figure}
\subsection{Sensitivity Analysis of Hyperparameters}
\label{subsec:sensitivity}

\begin{table}[t!]
\centering
\caption{Ablation study on RISS on DIOR-RSVG. ``Basic GRPO'' denotes the setting that uses only the target-only reward, while PAR denotes the proposed process-aware reward.}
\label{tab:dds_ablation}
\resizebox{\columnwidth}{!}{
\begin{tabular}{@{}l c ccc cc@{}}
\toprule
\multirow{2}{*}{\textbf{Setting}} & 
\multirow{2}{*}{\textbf{RISS}} & 
\multicolumn{3}{c}{\textbf{Outcome Metrics}} & 
\multicolumn{2}{c}{\textbf{Process Metrics}} \\
\cmidrule(lr){3-5} \cmidrule(l){6-7}
& & 
\textbf{Pr@0.5 (\%)} & 
\textbf{Pr@0.7 (\%)} & 
\textbf{mIoU} & 
\textbf{Faithful@0.5 $\uparrow$} & 
\textbf{Shortcut@0.5 $\downarrow$} \\
\midrule
Basic GRPO & \xmark & 80.52 & 69.41 & 70.75 & 51.03 & 15.42 \\
Basic GRPO & \cmark & 80.65 (\(\uparrow\)0.13) & 69.59 & 70.84 & 52.07 & 15.01 \\
\midrule
PAR & \xmark & 81.16 & 69.65 & 71.31 & 54.47 & 12.58 \\
\textbf{PAR} & \textbf{\cmark} & \textbf{81.52 (\(\uparrow\)0.36)} & \textbf{70.12} & \textbf{71.53} & \textbf{56.38} & \textbf{11.31} \\
\bottomrule
\end{tabular}
}
\end{table}
Figure~\ref{fig:alpha_curve} shows the sensitivity of the weight ratio between the target reward ($\alpha_{tgt}$) and the reference reward ($\alpha_{ref}$) in the PAR. The overall trend exhibits an inverted-U shape, with the best performance achieved around $[0.7,\,0.3]$, indicating that effective reasoning-oriented optimization requires a proper balance between final target localization and intermediate reference grounding. When $\alpha_{ref}$ is too low, optimization becomes overly dominated by the final target objective, weakening supervision on the intermediate reasoning process and making shortcut behavior more likely. In contrast, when $\alpha_{ref}$ is too high, the model tends to over-emphasize the intermediate anchor at the expense of final target localization.

Figure~\ref{fig:rids_threshold} analyzes the sensitivity of the RISS threshold. 
We report the relative performance change on DIOR-RSVG using $\tau_{\mathrm{RISS}}=0.4$ as the reference. 
The model achieves the best performance when $\tau_{\mathrm{RISS}}$ is set to 0.4, while either a lower or a higher threshold leads to performance degradation. 
This result is consistent with the design of RISS, since either an overly high or overly low threshold may select samples with saturated PAR responses, providing limited incremental supervision or weak discrimination among rollout trajectories.
Table~\ref{tab:aux_hyper} reports the sensitivity of auxiliary coefficients in PAR and the format reward.
The performance varies only slightly when $\lambda_{\mathrm{pen}}$, $\beta$, and $w_{\mathrm{fmt}}$ are changed from 0.1 to 0.3.
This indicates that GeoSearcher is not sensitive to these auxiliary coefficients.
We therefore use 0.2 as the default setting for all three coefficients.

\begin{table}[t]
\centering
\caption{Sensitivity analysis of auxiliary coefficients in PAR and format reward on DIOR-RSVG.
Here, $\lambda_{\mathrm{pen}}$ denotes the shared setting of $\lambda_{\mathrm{hal}}$ and $\lambda_{\mathrm{multi}}$.}
\label{tab:aux_hyper}
\resizebox{\columnwidth}{!}{
\begin{tabular}{@{}cc cc cc@{}}
\toprule
\multicolumn{2}{c}{\textbf{Penalty Coefficient}} &
\multicolumn{2}{c}{\textbf{Bonus Coefficient}} &
\multicolumn{2}{c}{\textbf{Format Reward}} \\
\cmidrule(lr){1-2} \cmidrule(lr){3-4} \cmidrule(l){5-6}
$\lambda_{\mathrm{pen}}$ & \textbf{Pr@0.5 (\%)} &
$\beta$ & \textbf{Pr@0.5 (\%)} &
$w_{\mathrm{fmt}}$ & \textbf{Pr@0.5 (\%)} \\
\midrule
0.1 & 81.47 & 0.1 & 81.50 & 0.1 & 81.48 \\
\textbf{0.2} & 81.52 & \textbf{0.2} & 81.52 & \textbf{0.2} & 81.52 \\
0.3 & 81.53 & 0.3 & 81.49 & 0.3 & 81.50 \\
\bottomrule
\end{tabular}
}
\end{table}
\subsection{Effect of Reference Anchor Quality}
\label{subsec:anchor_quality}

\begin{figure}[!t]
\centering
\includegraphics[width=1\linewidth]{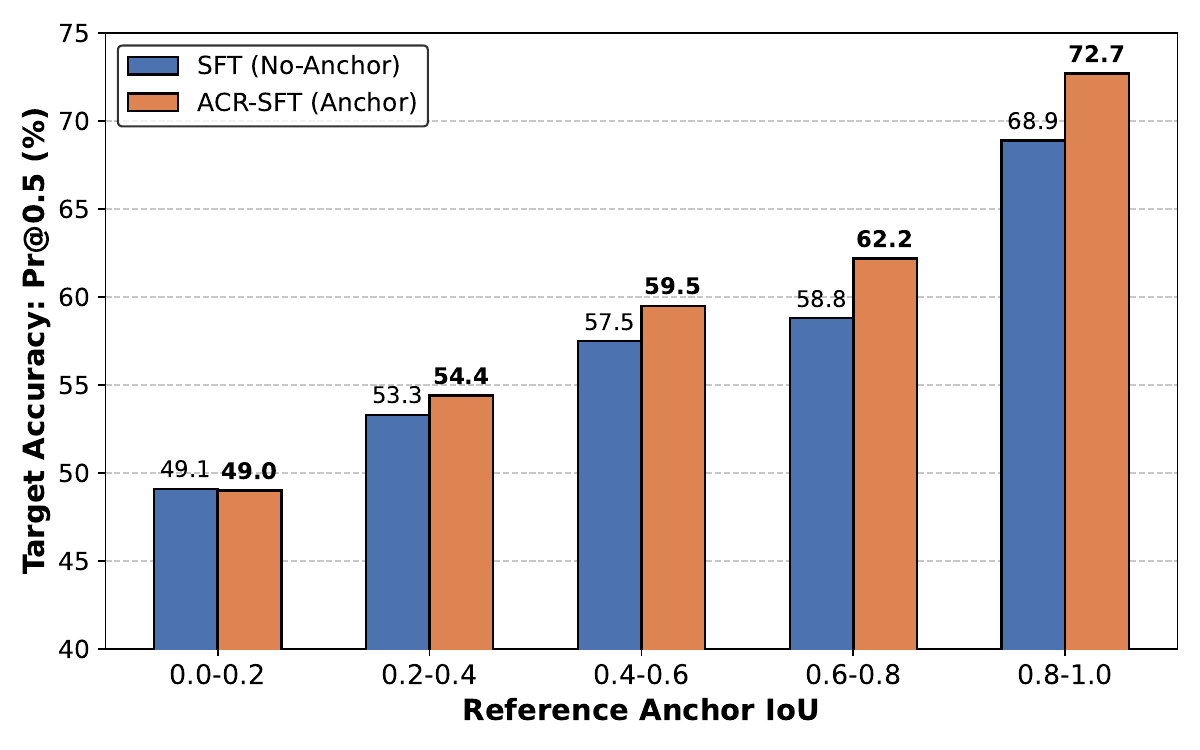}
\caption{Effect of reference anchor quality on final grounding accuracy. Test samples are grouped by the IoU between the predicted and ground truth reference boxes.}
\label{fig:ref_iou_bins}
% \vspace{-0.8em}
\end{figure}

To further analyze the role of reference anchor quality, we group the test samples according to the IoU between the reference anchors predicted by ACR-SFT and the ground truth reference coordinates. 
We then compare the final target localization performance of ACR-SFT with an SFT variant that does not insert explicit reference coordinates. 
Fig.~\ref{fig:ref_iou_bins} reports Pr@0.5 under different anchor quality intervals.
The results show a clear positive correlation between anchor quality and final grounding accuracy. 
As the anchor IoU increases, ACR-SFT achieves higher Pr@0.5, and its advantage over SFT becomes more evident. 
This indicates that a correctly grounded reference object can serve as an effective intermediate state. 
It helps the model narrow the search region before predicting the final target object.
However, the benefit of explicit anchors is not unconditional. 
When the reference anchor IoU is low, the improvement over SFT becomes marginal, and ACR-SFT may even perform slightly worse in the lowest-quality interval. 
This suggests that inaccurate reference grounding can mislead the subsequent reasoning process. 
In this case, the model may search for the target around an incorrect intermediate state.

\begin{figure}[!t]
\centering
\includegraphics[width=1\linewidth,trim=0 15 0 0, clip]{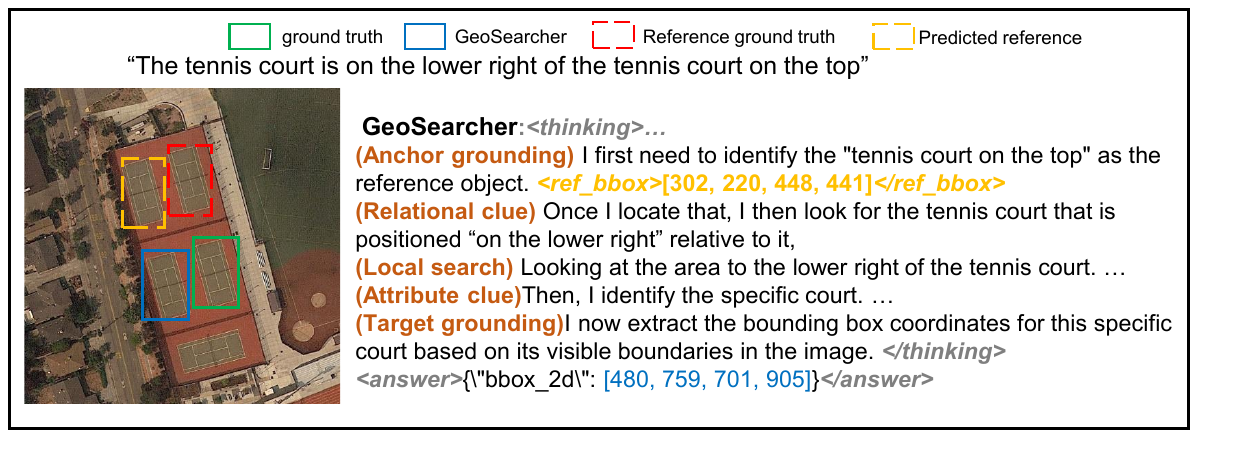}
\caption{A bad case caused by inaccurate reference anchor grounding. The predicted reference anchor (red dashed box) deviates from the corrected reference anchor (yellow dashed box), which shifts the subsequent relational search region and leads to an incorrect final prediction.}
\label{fig:bad_anchor_case}
\end{figure}

Fig.~\ref{fig:bad_anchor_case} shows a failure case where the model still follows the intended progressive reasoning pattern. 
It first predicts a reference anchor and then searches for the target according to the relational clue. 
Although the predicted reference anchor is not aligned with the corrected reference box, the subsequent prediction is still constrained to the local region indicated by the reference object, rather than drifting to arbitrary distractors in the full image. This suggests that anchor-guided reasoning can reduce the search burden compared with direct one-shot localization, which is more susceptible to visually similar distractors in large remote sensing scenes. The remaining error mainly comes from the inaccurate grounding of the reference object, which shifts the local search region and leads to a biased final prediction. This case further shows that accurate reference grounding is important for reliable progressive reasoning.

\section{Conclusion}

In this paper, we reformulate RSVG from one-shot coordinate generation into anchor-guided progressive reasoning, and propose GeoSearcher, an Anchor-Guided Progressive Reasoning Post-Training framework composed of ACR-SFT and PF-GRPO. In this framework, ACR-SFT injects a structured reasoning prior through explicit reference anchors, while PF-GRPO further aligns the intermediate reasoning process and final target localization with the PAR and RISS.
Experiments on DIOR-RSVG, OPT-RSVG, and VRS-Bench show that GeoSearcher achieves consistent improvements across multiple benchmarks. 

Future work: At the same time, our analysis reveals an important limitation of the current framework. The reference anchor quality study shows that the benefit of explicit anchor insertion strongly depends on whether the intermediate reference object is correctly grounded. When the reference anchor is accurate, ACR-SFT brings clear gains in final localization. Therefore, improving the robustness of reference grounding under more complex scenes, dense distractors, and weak visual clues therefore remains an important direction for future work.
In addition, the current GeoSearcher mainly models one primary reference anchor when reference-based reasoning is required. In real remote sensing queries, however, the target may be specified by multiple reference objects or by collaborative constraints among several landmarks. Future work will explore multi-reference collaborative reasoning, enabling the model to use multiple intermediate states more reliably under complex compositional relations and spatial layouts.

\bibliographystyle{IEEEtran}
\bibliography{referenceIEEE}

\newpage

\vfill

\end{document}